\documentclass[runningheads]{llncs}
\usepackage{graphicx}
\usepackage[acronym]{glossaries}
\usepackage[super]{nth}
\usepackage{booktabs}
\usepackage{xcolor}
\usepackage{pifont} % http://ctan.org/pkg/pifont
\usepackage{multirow}
\usepackage{notations}

\usepackage{tikz}
\usepackage{comment}
\usepackage{amsmath,amssymb} % define this before the line numbering.
\usepackage{color}

\usepackage[accsupp]{axessibility}  % Improves PDF readability for those with disabilities.

\usepackage[pagebackref=true,breaklinks=true,letterpaper=true,citecolor=blue,colorlinks,linkcolor=red,bookmarks=false]{hyperref}

\begin{document}

\pagestyle{headings}
\mainmatter
\def\ECCVSubNumber{5585}  % Insert your submission number here

\title{Robust Visual Tracking by Segmentation} % Replace with your title

\begin{comment}
\titlerunning{ECCV-22 submission ID \ECCVSubNumber} 
\authorrunning{ECCV-22 submission ID \ECCVSubNumber} 
\author{Anonymous ECCV submission}
\institute{Paper ID \ECCVSubNumber}
\end{comment}
%******************

% \begin{comment}
\title{Robust Visual Tracking by Segmentation}
\titlerunning{Robust Visual Tracking by Segmentation}

\author{Matthieu Paul \and Martin Danelljan \and Christoph Mayer \and Luc Van Gool\index{Van Gool, Luc}}
\authorrunning{Paul et al.}

\institute{Computer Vision Lab, ETH Z{\"u}rich, Switzerland\\
\email{\{paulma, damartin, chmayer, vangool\}@vision.ee.ethz.ch}}
% \end{comment}
%******************
\maketitle

\begin{abstract}

Estimating the target extent poses a fundamental challenge in visual object tracking. Typically, trackers are \textit{box-centric} and fully rely on a bounding box to define the target in the scene. In practice, objects often have complex shapes and are not aligned with the image axis. In these cases, bounding boxes do not provide an accurate description of the target and often contain a majority of background pixels.

We propose a \textit{segmentation-centric} tracking pipeline that not only produces a highly accurate segmentation mask, but also internally works with segmentation masks instead of bounding boxes. Thus, our tracker is able to better learn a target representation that clearly differentiates the target in the scene from background content. In order to achieve the necessary robustness for the challenging tracking scenario, we propose a separate instance localization component that is used to condition the segmentation decoder when producing the output mask. We infer a bounding box from the segmentation mask, validate our tracker on challenging tracking datasets and achieve the new state of the art on LaSOT with a success \acrshort{auc} score of 69.7\%. Since most tracking datasets do not contain mask annotations, we cannot use them to evaluate predicted segmentation masks. Instead, we validate our segmentation quality on two popular video object segmentation datasets. The code and trained models are available at \url{https://github.com/visionml/pytracking}.

\end{abstract}

%-------------------------------------------------------------------------
\section{Introduction}

\begin{figure*}[t]
\centering
\includegraphics[width=1.0\linewidth]{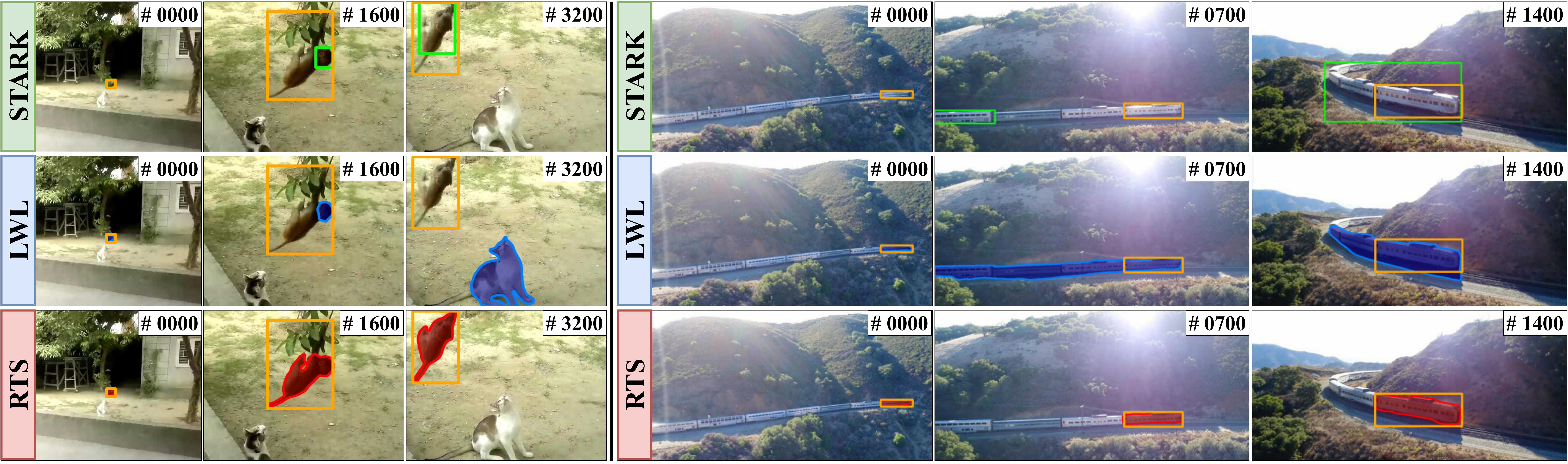}
\caption{Comparison between the \gls{vot} method Stark~\cite{Yan_2021_ICCV_STARK}, the \gls{vos} method LWL~\cite{Bhat_2020_ECCV_LWL} and our proposed method on two tracking sequences from the LaSOT~\cite{LaSOT} dataset. The ground-truth annotation (\bbox{_orange}) is shown in each frame for reference. Our approach is more robust and predicts a more accurate target representation.}
\label{fig:teaser_mosaic}
\end{figure*}

Visual object tracking is the task of estimating the state of a target object for each frame in a video sequence. The target is solely characterized by its initial state in the video. Current approaches predominately characterize the state itself with a bounding box. However, this only gives a very coarse representation of the target in the image. In practice, objects often have complex shapes, undergo substantial deformations. Often, targets do not align well with the image axes, while most benchmarks use axis-aligned bounding boxes. In such cases, the majority of the image content inside the target's bounding box often consists of background regions which provide limited information about the object itself. In contrast, a segmentation mask precisely indicates the object's extent in the image (see Fig.~\ref{fig:teaser_mosaic} frames \#1600 and \#3200).  Such information is vital in a variety of applications, including video analysis, video editing, and robotics. 
In this work, we therefore develop an approach for accurate and robust target object segmentation, even in the highly challenging tracking datasets~\cite{LaSOT,Mueller_2016_ECCV_UAV123}.

While severely limiting the information about the target's state in the video, the aforementioned issues with the bounding box representation can itself lead to inaccurate bounding box predictions, or even tracking failure.
To illustrate this, Fig.~\ref{fig:teaser_mosaic} shows two typical tracking sequences. The tracking method STARK~\cite{Yan_2021_ICCV_STARK} (first row) fails to regress bounding boxes that contain the entire object (\#1600, \#1400) or even starts tracking the wrong object (\#0700). Conversely, segmentation masks are a better fit to differentiate pixels in the scene that belong to the background and the target. Therefore, a \textit{segmentation-centric} tracking architecture designed to work internally with a segmentation mask of the target instead of a bounding box has the potential to learn better target representations, because it can clearly differentiate background from foreground regions in the scene.

A few recent tracking methods~\cite{Voigtlaender_2020_CVPR_SiamRCNN,Yan2021AlphaRefineBT} have recognized the advantage of producing segmentation masks instead of bounding boxes as final output. However, these trackers are typically \textit{bounding-box-centric} and the final segmentation mask is obtained by a separate \textit{box-to-mask} post-processing network. These methods do not leverage the accurate target definition of segmentation masks to learn a more accurate and robust internal representation of the target. 

In contrast, most Video Object Segmentation (VOS) methods~\cite{Oh_2019_ICCV_STM,Bhat_2020_ECCV_LWL} follow a \textit{segmentation-centric} paradigm. However, these methods are not designed for the challenging tracking scenarios. Typical VOS sequences consist only of a few hundred frames~\cite{Pont-Tuset_arXiv_2017_DAVIS} whereas multiple sequences of more than ten thousand frames exist in tracking datasets~\cite{LaSOT}. Due to this setup, VOS methods focus on producing highly accurate segmentation masks but are sensitive to distractors, substantial deformations and occlusions of the target object. Fig.~\ref{fig:teaser_mosaic} shows two typical tracking sequences where the VOS method LWL~\cite{Bhat_2020_ECCV_LWL} (second row) produces a fine-grained segmentation mask of the wrong object (\#3200) or is unable to detect only the target within a crowd (\#0700, \#1400). 

We propose \emph{Robust Visual Tracking by Segmentation} (RTS), a unified tracking architecture capable of predicting accurate segmentation masks. To design a \textit{segmentation-centric} approach, we take inspiration from the aforementioned LWL~\cite{Bhat_2020_ECCV_LWL} method. However, to achieve robust and accurate segmentation on Visual Object Tracking (VOT) datasets, we introduce several new components. In particular, we propose an instance localization branch trained to predict a target appearance model, which allows occlusion detection and target identification even in cluttered scenes. The output of the instance localization branch is further used to condition the high-dimensional mask encoding. This allows the segmentation decoder to focus on the localized target, leading to a more robust mask prediction. Since our proposed method contains a segmentation and instance memory that need to be updated with previous tracking results, we design a memory management module. This module first assesses the prediction quality, decides whether the sample should enter the memory and, when necessary, triggers the model update.

\parsection{Contributions} Our contributions are the following:
\textbf{(i)} We propose a unified tracking architecture capable of predicting robust classification scores and accurate segmentation masks. We design separate feature spaces and memories to ensure optimal receptive fields and update rates for segmentation and instance localization.
\textbf{(ii)}  To produce a segmentation mask which agrees with the instance prediction, we design a fusion mechanism that conditions the segmentation decoder on the instance localization output and leads to more robust tracking performance.
\textbf{(iii)} We introduce an effective inference procedure capable of fusing the instance localization output and mask encoding to ensure both robust and accurate tracking.
\textbf{(iv)} We perform comprehensive evaluation and ablation studies of the proposed tracking pipeline on multiple popular tracking benchmarks. Our approach achieves the new state of the art on LaSOT with an \gls{auc} score of 69.7\%.

%-------------------------------------------------------------------------
\section{Related Work}

\parsection{Visual Object Tracking}

Over the years, research in the field of visual tracking has been accelerated by the introduction of new and challenging benchmarks, such as LaSOT~\cite{LaSOT}, GOT-10k~\cite{GOT10k}, and TrackingNet~\cite{TrackingNet}.
This led to the introduction of new paradigms in visual object tracking, based on \glspl{dcf}, Siamese networks and Transformers.

One of the most popular type of approaches, DCF-based visual trackers \cite{Bolme_2010_CVPR_MOSSE,Henriques_2015_TPAMI_KCF,Danelljan_2016_ECCV_CCOT,Lukezic_2018_IJCV_CSRDCF,danelljan_CVPR_2019_ATOM,Wang_2020_CVPR_MAML,Zheng_2020_ECCV_DCFST,bhat_ICCV_2019_DiMP,danelljan_ICCV_2019_PRDimP} essentially solve an optimization problem to estimate the weights of the \gls{dcf} that allow to distinguish foreground from background regions. The \gls{dcf} is often referred to as the target appearance model and allows to localize the target in the video frame. More recent \gls{dcf} approaches~\cite{bhat_ICCV_2019_DiMP,danelljan_ICCV_2019_PRDimP} enable end-to-end training by unrolling a fixed number of the optimization iterations during \emph{offline} training.

Siamese tracking methods have gained in popularity due to their simplicity, speed and end-to-end trainability~\cite{Valmadre_2017_CVPR_Siamese,Bertinetto_2016_ECCVW_SiameseFC,Tao2016Sint,DaSiamRPN,DSiam,RASNet,SASiamR,Li_2018_CVPR_SiamRPN,Li_2019_CVPR_SiamRPN++}. These trackers learn a similarity metric using only the initial video frame and its annotation that allows to clearly identify the target \emph{offline}. Since no \emph{online} learning component is involved, these trackers achieve high frame rates at the cost of limited \emph{online} adaptability to changes of the target's appearance. Nonetheless, several methods have been proposed to overcome these issues~\cite{Valmadre_2017_CVPR_Siamese,Bertinetto_2016_ECCVW_SiameseFC,Li_2018_CVPR_SiamRPN,Li_2019_CVPR_SiamRPN++}.

Very recently, Transformer-based trackers have achieved state-of-the-art performance on many datasets, often outperforming their rivals. This group of trackers typically uses a Transformer component in order to fuse information extracted from training and test frames. This produces discriminative features that allow to accurately localize and estimate the target in the scene~\cite{Chen_2021_CVPR_TransT,Yu_2021_ICCV_HPF,Yan_2021_ICCV_STARK,Wang_2021_CVPR_TrDiMP,Mayer_2022_CVPR_ToMP}.

\parsection{Video Object Segmentation}
Semi-supervised \gls{vos} is the task of classifying all pixels belonging to the target in each video frame, given only the segmentation mask of the target in the initial frame. The cost of annotating accurate segmentation masks is limiting the sequence length and number of videos contained in available VOS datasets. Despite the relatively small size of \gls{vos} datasets compared to other computer vision problems, new benchmarks such as Youtube-VOS~\cite{xu2018youtubevos} and DAVIS~\cite{Pont-Tuset_arXiv_2017_DAVIS} accelerated the research progress in the last years. 

Some methods rely on a learnt target detector \cite{caelles2017_OSVOS,voigtlaender2017online,maninis2018video}, others learn how to propagate the segmentation mask across frames~\cite{wug2018fast,perazzi2017learning,li2018video,khoreva2017lucid}. Another group of methods uses feature matching techniques across one or multiple frames with or without using an explicit spatio-temporal memory~\cite{chen2018blazingly,hu2018videomatch,voigtlaender2019feelvos,Oh_2019_ICCV_STM}. Recently, Bhat~\etal~\cite{Bhat_2020_ECCV_LWL} employed meta-learning approach, introducing an end-to-end trainable \gls{vos} architecture. In this approach, a few-shot learner predicts a learnable labels encoding. It generates and updates \emph{online} the parameters of a segmentation target model that produces the mask encoding used to generate the final segmentation mask.

\parsection{Joint Visual Tracking and Segmentation}
A group of tracking methods have already identified the advantages of predicting a segmentation mask instead of a bounding box~\cite{Yan2021AlphaRefineBT,zhao_ICCV_2021_vidboxseg_STA,Voigtlaender_2020_CVPR_SiamRCNN,Lukezic_CVPR_2020_D3S,wang2019fast_SiamMask,Son_2015_ICCV_trackbyseg}.
Siam-RCNN is a box-centric tracker that uses a pretrained \emph{box2seg} network to predict the segmentation mask given a bounding box prediction. In contrast, AlphaRefine represents a novel \emph{box2seg} method that has been evaluated with many recent trackers such as SuperDiMP~\cite{danelljan_ICCV_2019_PRDimP} and SiamRPN++~\cite{Li_2019_CVPR_SiamRPN++}. Further, Zhao~\etal~\cite{zhao_ICCV_2021_vidboxseg_STA} focus on generating segmentation masks from bounding box annotations in videos using a spatio-temporal aggregation module to mine consistencies of the scene across multiple frames.
Conversely, SiamMask~\cite{wang2019fast_SiamMask} and D3S~\cite{Lukezic_CVPR_2020_D3S} are segmentation-centric trackers that produce a segmentation mask directly, without employing a \emph{box2seg} module. In particular, SiamMask~\cite{wang2019fast_SiamMask} is a fully-convolutional Siamese network with a separate branch which predicts binary segmentation masks supervised by a segmentation loss.

From a high-level view, the single-shot segmentation tracker D3S~\cite{Lukezic_CVPR_2020_D3S} is most related to our proposed method. Both methods employ two dedicated modules or branches; one for localization and one for segmentation. D3S adopts the target classification component of ATOM~\cite{danelljan_CVPR_2019_ATOM}, requiring online optimization of weights in a two-layer CNN. In contrast, we learn online the weights of a \gls{dcf} similar to DiMP~\cite{bhat_ICCV_2019_DiMP}. For segmentation, D3S~\cite{Lukezic_CVPR_2020_D3S} propose a feature matching technique that matches test frame features with background and foreground features corresponding to the initial frame. In contrast, we adopt the few-shot learning based model prediction proposed in LWL~\cite{Bhat_2020_ECCV_LWL} to produce accurate segmentation masks. Furthermore, D3S proposes to simply concatenate the outputs of both modules whereas we learn a localization encoding to condition the segmentation mask decoding based on the localization information. Compared to D3S, we update not only the instance localization but also the segmentation models and memories. Hence, our method integrates specific memory management components. 

%-------------------------------------------------------------------------
\begin{figure*}[t]
\centering
\includegraphics[width=1.0\linewidth]{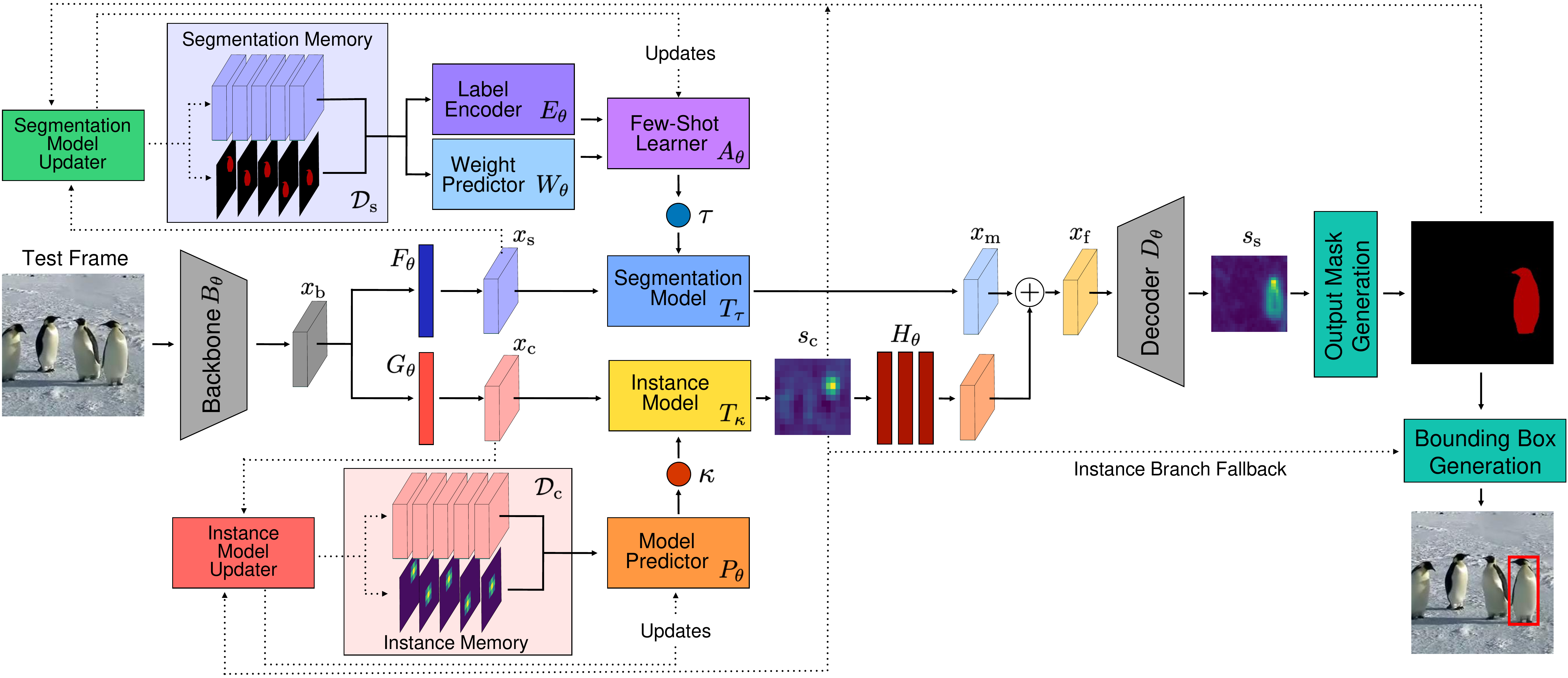}
\caption{Overview of our entire online tracking pipeline used for inference, see Sec~\ref{sec:method_overview}.}
\label{fig:overview}
\end{figure*}

\section{Method}

\subsection{Overview}
\label{sec:method_overview}
Video object segmentation methods can produce high quality segmentation masks but are typically not robust enough for video object tracking. Robustness becomes vital for medium and long sequences, which are most prevalent in tracking datasets~\cite{LaSOT,Mueller_2016_ECCV_UAV123}. In such scenarios, the target object frequently undergoes substantial appearance changes. Occlusions and similarly looking objects are common. Hence, we propose to adapt a typical \gls{vos} approach with tracking components to increase its robustness. In particular, we base our approach on the \gls{lwl}~\cite{Bhat_2020_ECCV_LWL} method and design a novel and segmentation-centric tracking pipeline that estimates accurate object masks instead of bounding boxes. During inference, a segmentation mask is typically not provided in visual object tracking. Hence, we use STA~\cite{zhao_ICCV_2021_vidboxseg_STA} to generate a segmentation mask from the provided initial bounding box. 
An overview of our RTS method is shown in Fig.~\ref{fig:overview}. Our pipeline consists of a backbone network, a segmentation branch, an instance localization branch and a segmentation decoder. For each video frame, the backbone first extracts a feature map $\bboneFeat$. These features are further processed into segmentation features $\segFeat$ and classification features $\clfFeat$ to serve as input for their respective branch. The segmentation branch is designed to capture the details of the object with a high dimensional mask encoding, whereas the instance localization branch aims at providing a coarser but robust score map representing the target location. Both branches contain components learned online, trained on memories ($\segSupportSet$ and $\clfSupportSet$) that store features and predictions of past frames. The instance localization branch has two purposes. The first is to control models and memories updating. The second is used to condition the segmentation mask decoder. To do so, we add instance localization information with a learnt score encoding produced by $\clfScoresEnc$. The obtained segmentation scores and the raw instance model score map are then used to generate the final segmentation mask output.

\subsection{Segmentation Branch}
\label{sec:seg_branch}

The architecture of the segmentation branch is adopted from LWL~\cite{Bhat_2020_ECCV_LWL}, and we briefly review it here. It consists of a segmentation sample memory $\segSupportSet$, a label generator $\labelGenerator$, a weight predictor $\weightpredictor$, a few-shot learner $\fewshotlearner$ and a segmentation model $\segTargetModel$. The goal of the few-shot learner $\fewshotlearner$ is producing the parameters $\segparam$ of the segmentation model $\segTargetModel$ such that the obtained mask encoding $\maskencoding$ contains the information needed to compute the final segmentation mask of the target object. The label mask encodings used by the few-shot learner are predicted by the label generator $\labelGenerator$. 

The few-shot learner is formulated through the following optimization problem, which is unrolled through steepest descent iterations in the network 

\begin{equation}
\label{eq:total_seg_loss}
    \totalSegLoss(\segparam) = \frac{1}{2}\sum_{(\segFeat, \segLabel) \in \segSupportSet} \big\| \weightpredictor(\segLabel) \cdot \big(\segTargetModel(\segFeat) - \labelGenerator(\segLabel)\big) \big\|^2 + \frac{\lambda_\text{s}}{2}\|\segparam\|^2,
\end{equation}
where $\segSupportSet$ corresponds to the segmentation memory, $x_s$ denotes the segmentation features, $y_s$ the segmentation masks and $\lambda_\text{s}$ is a learnable scalar regularization parameter. The weight predictor $\weightpredictor$ produces sample confidence weights for each spatial location in each memory sample. 
Applying the optimized model parameters $\optsegparam$ within the segmentation model produces the mask encoding $\maskencoding = \segoptTargetModel(\segFeat)$ for the segmentation features $\segFeat$.

LWL~\cite{Bhat_2020_ECCV_LWL} feeds the mask encoding directly into the segmentation decoder to produce the segmentation mask. For long and challenging tracking sequences, only relying on the mask encoding may lead to an accurate segmentation mask, but often for the wrong object in the scene (see Fig~\ref{fig:teaser_mosaic}). Since LWL~\cite{Bhat_2020_ECCV_LWL} is only able to identify the target to a certain degree in challenging tracking sequences, we propose to condition the mask encoding based on an instance localization representation, described next.

\subsection{Instance Localization Branch}
\label{sec:instance_branch}

The segmentation branch can produce accurate masks but typically lacks the necessary robustness for tracking in medium or long-term sequences. Especially challenging are sequences where objects similar to the target appear, where the target object is occluded or vanishes from the scene for a short time. Therefore, we propose a dedicated branch for target instance localization, in order to robustly identify the target  among distractors or to detect occlusions. A powerful tracking paradigm that learns a target-specific appearance model on both foreground and background information are discriminative correlation filters (DCF)~\cite{Bolme_2010_CVPR_MOSSE,Henriques_2015_TPAMI_KCF,Danelljan_2017_CVPR_ECO,bhat_ICCV_2019_DiMP}. These methods learn the weights of a filter that differentiates foreground from background pixels represented by a score map, where the maximal value corresponds to the target's center.  

Similar to the segmentation branch, we propose an instance localization branch that consists of a sample memory $\clfSupportSet$ and a model predictor $\clfModelPredictor$. The latter predicts the parameters $\clfparam$ of the instance model $\clfTargetModel$. The instance model is trained online to produce the target score map used to localize the target object. To obtain the instance model parameters $\clfparam$ we minimize the following loss function
\begin{equation}\label{eq:classification-loss}
    \totalClfLoss(\clfparam) = \sum_{(\clfFeat, \clfLabel) \in \clfSupportSet} \left\|\internalClfloss\big(\clfTargetModel(\clfFeat), \clfLabel\big)\right\|^2 + \frac{\lambda_\text{c}}{2}\|\clfparam\|^2,
\end{equation}
where $\clfSupportSet$ corresponds to the instance memory containing the classification features $\clfFeat$ and the Gaussian labels $\clfLabel$. $\internalClfloss$ denotes the robust hinge-like loss~\cite{bhat_ICCV_2019_DiMP} and $\lambda_\text{c}$ is a fixed regularization parameter. To solve the optimization problem we apply the method from \cite{bhat_ICCV_2019_DiMP}, which unrolls steepest descent iterations of the Gauss-Newton approximation of \eqref{eq:classification-loss} to obtain the final model parameters $\optclfparam$. The score map can then be obtained with $\clfscores = \clfoptTargetModel(\clfFeat)$ by evaluating the target model on the classification features $\clfFeat$.

\subsection{Instance-Conditional Segmentation Decoder}
In video object segmentation the produced mask encoding is directly fed into the segmentation decoder to generate the segmentation mask. However, solely relying on the mask encoding is not robust enough for the challenging tracking scenario, see Fig~\ref{fig:teaser_mosaic}. Thus, we propose to integrate the instance localization information into the segmentation decoding procedure. In particular, we condition the mask encoding on a learned encoding of the instance localization score map. 

First, we encode the raw score maps using a multi-layer \gls{cnn} to learn a suitable representation. Secondly, we condition the mask encoding with the learned representation using element-wise addition. The entire conditioning procedure can be defined as $x_\text{f} = \maskencoding + \clfScoresEnc(\clfscores)$, where $\clfScoresEnc$ denotes the \gls{cnn} encoding the scores $\clfscores$, and $\maskencoding$ the mask encoding. The resulting features are then fed into the segmentation decoder that produces the segmentation scores of the target object.

\subsection{Jointly Learning Instance Localization and Segmentation}\label{sec:training}

In this section, we describe our general training strategy and parameters. In particular, we further detail the segmentation and classification losses that we use for offline training. 

\parsection{Segmentation Loss} First, we randomly sample $J$ frames from an annotated video sequence and sort them according to their frame IDs in increasing order to construct the training sequence $\mathcal{V} = \{(\bboneFeat^j, \segLabel^j, \clfLabel^j)\}_{j=0}^{J-1}$, where $\bboneFeat^j = \bbone(I^j)$ are the extracted features of the video frame $I^j$ using the backbone $\bbone$, $\segLabel^j$ is the corresponding segmentation mask and $\clfLabel^j$ denotes the Gaussian label at the target's center location. We start with entry $v_0 \in \mathcal{V}$ and store it in the segmentation $\segSupportSet$ and instance memory $\clfSupportSet$ and obtain parameters $\segparam^0$ and $\clfparam^0$ of the segmentation and instance model. We use these parameters to compute the segmentation loss for $v_1\in \mathcal{V}$. Using the predicted segmentation mask, we update the segmentation model parameters to $\segparam^1$ but keep the instance model parameters fixed. Segmentation parameters typically need to be updated frequently to enable accurate segmentation. Conversely, we train the model predictor on a single frame only. The resulting instance model generalizes to multiple unseen future frames, ensuring robust target localization. The resulting segmentation loss for the entire sequence $\mathcal{V}$ can thus be described as follows
\begin{equation}
    \trainSegLoss(\theta; \mathcal{V}) = \sum_{j=1}^{J-1}\mathcal{L}_\text{s}\bigg(D_\netparam\Big(T_{\segparam^{j-1}}(\segFeat^j) + \clfScoresEnc\big(T_{\clfparam^{0}}(\clfFeat^j)\big)\Big),~\segLabel^j\bigg),
\label{eq:train_segmentation_loss}
\end{equation}
where $\segFeat = \segExtractor(\bboneFeat)$ and $\clfFeat = \clfExtractor(\bboneFeat)$ and $\mathcal{L}_\text{s}$ is the Lovasz segmentation loss~\cite{Berman_2018_CVPR_lovasz}.

\parsection{Classification Loss} Instead of training our tracker only with the segmentation loss, we add an auxiliary loss to ensure that the instance module produces score maps localizing the target via a Gaussian distribution. These score maps are essential to update the segmentation and instance memories and to generate the final output.
As explained before, we use only the first training $v_0 \in \mathcal{V}$ to optimize the instance model parameters. To encourage fast convergence, we use not only the parameters corresponding to the final iteration $N_\mathrm{iter}$ of the optimization method $\clfparam^{0}_{(N_\mathrm{iter})}$ explained in Sec.~\ref{sec:instance_branch}, but also all the intermediate parameters $\clfparam^{0}_{(i)}$ of loss computation. The final target classification loss for the whole sequence $\mathcal{V}$ is defined as follows
\begin{equation}
    \trainClfLoss(\theta; \mathcal{V}) = \sum_{j=1}^{J-1}\left(\frac{1}{N_\text{iter}}\sum_{i=0}^{N_\text{iter}}\mathcal{L}_\text{c}\Big(T_{\clfparam^{0}_{(i)}}(\clfFeat^j),~ \clfLabel^j\Big)\right),
\label{eq:train_classification_loss}
\end{equation}
where $\mathcal{L}_\text{c}$ is the hinge loss defined in~\cite{bhat_ICCV_2019_DiMP}.
To train our tracker we combine the segmentation and classification losses using the scalar weight $\eta$ and minimize both losses jointly
\begin{equation}
    \mathcal{L}_\text{tot}^\text{seq}(\theta; \mathcal{V}) =  \mathcal{L}_\text{s}^\text{seq}(\theta; \mathcal{V}) + \eta \cdot\mathcal{L}_\text{c}^\text{seq}(\theta; \mathcal{V}).
\label{eq:train_combined_loss}
\end{equation}

\parsection{Training Details}
We use the train sets of LaSOT~\cite{LaSOT}, GOT-10k~\cite{GOT10k}, Youtube-VOS~\cite{xu2018youtubevos} and DAVIS~\cite{Pont-Tuset_arXiv_2017_DAVIS}. For \gls{vot} datasets that only provide annotated bounding boxes, we use these boxes and STA~\cite{zhao_ICCV_2021_vidboxseg_STA} to generate segmentation masks and treat them as ground truth annotations during training. STA~\cite{zhao_ICCV_2021_vidboxseg_STA} is trained separately on YouTube-VOS 2019~\cite{xu2018youtubevos} and DAVIS 2017~\cite{Perazzi2016}.
For our model, we use ResNet-50 with pre-trained MaskRCNN weights as our backbone and initialize the segmentation model and decoder weights with the ones available from LWL~\cite{Bhat_2020_ECCV_LWL}. We train for 200 epochs and sample 15'000 videos per epoch, which takes 96 hours to train on a single Nvidia A100 GPU. We use the ADAM~\cite{Kingma_2014_ADAM} optimizer with a learning rate decay of 0.2 at epochs 25, 115 and 160. We weigh the losses such that the segmentation loss is predominant but in the same range as the classification loss. We empirically choose $\eta=10$. Further details about training and the network architecture are given in the appendix.

\subsection{Inference}

\parsection{Memory Management and Model Updating}
Our tracker consists of two different memory modules. A segmentation memory that stores segmentation features and predicted segmentation masks of previous frames. In contrast, an instance memory contains classification features and Gaussian labels marking the center location of the target in the predicted segmentation mask of the previous video frame. The quality of the predicted labels directly influences the localization and segmentation quality in future video frames. Hence, it is crucial to avoid contaminating the memory modules with predictions that do not correspond to the actual target. We propose the following strategy to keep the memory as clean as possible. (a) If the instance model is able to clearly localize the target (maximum value in the score map larger than $t_{s_c} = 0.3$) and the segmentation model constructs a valid segmentation mask (at least one pixel above $t_{s_s}=0.5$) we update both memories with the current predictions and features. (b) If either the instance localization or segmentation fail to identify the target we omit updating the segmentation memory. (c) If only the segmentation mask fails to represent the target but the instance model can localize it, we update the instance memory only. (d) If instance localization fails we do not update either memory. Further, we trigger the few-shot learner and model predictor after 20 frames have passed, but only if the corresponding memory has been updated.

\parsection{Final Mask Output Generation}
We obtain the final segmentation mask by thresholding the segmentation decoder output. To obtain the bounding box required for standard tracking benchmarks, we report the smallest axis-aligned box that contains the entire estimated object mask. 

\parsection{Inference Details} 
We set the input image resolution such that the segmentation learner features have a resolution of $52\times30$ (stride 16), while the instance learner operates on features of size $26\times15$ (stride 32). The learning rate is set to 0.1 and 0.01 for the segmentation and instance learner respectively. We use a maximum buffer of 32 frames for the segmentation memory and 50 frames for the instance memory. We keep the samples corresponding to the initial frame in both memories and replace the oldest entries if the memory is full. We update both memories for the first 100 video frames and afterwards only after every \nth{20} frame.
We randomly augment the sample corresponding to the initial frame with vertical flip, random translation and blurring.

%-------------------------------------------------------------------------
\section{Evaluation}
\label{sec:evaluation}

Our approach is developed within the PyTracking~\cite{pytracking} framework. The implementation is done with PyTorch 1.9 with CUDA 11.1. Our model is evaluated on a single Nvidia GTX 2080Ti GPU. Our method achieves an average speed of 30 FPS on LaSOT~\cite{LaSOT}. Each number corresponds to the average of five runs with different random seeds.

\begin{table}[!b]
    \caption{Comparison between our segmentation network baseline LWL and our pipeline, with and without Instance conditioning on different VOT datasets.}
    \centering
    \newcommand{\dist}{\quad}%
    \resizebox{1.0\columnwidth}{!}{%
    \begin{tabular}{l@{\dist}|@{\dist}c@{\dist}c@{\dist}|@{\dist}c@{\dist}c@{\dist}c@{\dist}|@{\dist}c@{\dist}c@{\dist}c@{\dist}|@{\dist}c@{\dist}c@{\dist}c@{\dist}c@{\dist}c@{\dist}}
        \toprule
        \multirow{2}{*}{Method}            &  Seg.    &  Inst. Branch & \multicolumn{3}{c}{LaSOT~\cite{LaSOT}}   & \multicolumn{3}{c}{NFS~\cite{Galoogahi_2017_ICCV_NFS}} & \multicolumn{3}{c}{UAV123~\cite{Mueller_2016_ECCV_UAV123}} \\
                    &  Branch  &  Conditioning & AUC    & P       & NP        & AUC    & P       & NP   & AUC    & P       & NP             \\ 
    \midrule
    LWL~\cite{Bhat_2020_ECCV_LWL} &   \cmark &  -          & 59.7   & 60.6    &  63.3     & 61.5  & 75.1  & 76.9  & 59.7  & 78.8 & 71.4 \\
    \textbf{RTS}           &   \cmark &  \xmark     & 65.3   & 68.5    &  71.5     & 65.8  & 84.0  & 85.0  & 65.2  & 85.6 & 78.8 \\
    \textbf{RTS}                 &   \cmark &  \cmark     & 69.7   & 73.7    &  76.2     & 65.4  & 82.8  & 84.0  & 67.6  & 89.4 & 81.6 \\\bottomrule
    
    \end{tabular}}
    \label{tab:ablation_branches}
\end{table}

\subsection{Branch Ablation Study}
\label{sec:ablation}

For the ablation study, we analyze the impact of the instance branch on three datasets and present the results in Tab.~\ref{tab:ablation_branches}. First, we report the performance of LWL~\cite{Bhat_2020_ECCV_LWL} since we build upon it to design our final tracking pipeline. We use the network weights provided by Bhat~\etal~\cite{Bhat_2020_ECCV_LWL} and the corresponding inference settings. We input the same segmentation masks obtained from the initial bounding box for LWL as used for our method. We observe that LWL is not robust enough for challenging tracking scenarios. The second row in Tab.~\ref{tab:ablation_branches} corresponds to our method but we omit the proposed instance branch. Hence, we use the proposed inference components and settings and train the tracker as explained in Sec.~\ref{sec:training}, but with conditioning removed. We observe that even without the instance localization branch our tracker can achieve competitive performance on all three datasets (\eg~+5.6\% on LaSOT). Fully integrating the instance localization branch increases the performance even more (\eg~+4.4 on LaSOT). Thus, we conclude that adapting the baseline method to the tracking domain improves the tracking performance. To boost the performance and achieve state-of-the-art results, an additional component able to increase the tracking robustness is required.

\subsection{Inference Parameters}
\label{sec:inference_params}
\begin{table}[!t]
    \caption{Ablation on inference strategies. The first column analyzes the effect of using the instance branch as fallback for target localization if the segmentation branch is unable to detect the target ($\max(s_s) < t_{s_s}$). The second column shows the impact of different confidence thresholds $t_{s_c}$.}
    \centering
    \newcommand{\dist}{\quad}%
    
    \resizebox{1.0\columnwidth}{!}{%
        \begin{tabular}{c@{\dist}c@{\dist}|@{\dist}c@{\dist}c@{\dist}c@{\dist}|@{\dist}c@{\dist}c@{\dist}c@{\dist}|@{\dist}c@{\dist}c@{\dist}c@{\dist}}
        \toprule
        
        Inst. Branch   &  \multirow{2}{*}{$t_{s_c}$}          & \multicolumn{3}{c}{LaSOT~\cite{LaSOT}}  & \multicolumn{3}{c}{NFS~\cite{Galoogahi_2017_ICCV_NFS}}    & \multicolumn{3}{c}{UAV123~\cite{Mueller_2016_ECCV_UAV123}} \\
        Fallback       &   & \gls{auc}    & P       & NP & \gls{auc}    & P       & NP & \gls{auc}    & P       & NP \\ 
        \midrule
        \xmark       &  0.30       &  69.3  & 73.1    & 75.9  & 65.3  & 82.7  & 84.0  & 66.3  & 87.2 & 80.4 \\
        \cmark       &  0.30       &  69.7  & 73.7    & 76.2  & 65.4  & 82.8  & 84.0  & 67.6  & 89.4 & 81.6 \\

        \midrule
        \midrule
        \cmark       &  0.20       &  68.6  & 72.3    & 75.0  & 65.3  & 82.7  & 83.9  & 67.0  & 88.7 & 80.7 \\
        \cmark       &  0.30       &  69.7  & 73.7    & 76.2  & 65.4  & 82.8  & 84.0  & 67.6  & 89.4 & 81.6 \\
        \cmark       &  0.40       &  69.1  & 72.7    & 75.6  & 63.3  & 79.7  & 81.7  & 67.1  & 89.1 & 80.7\\ \bottomrule

    \end{tabular}}
    \label{tab:ablation_strategies}
\end{table}

In this part, we ablate two key aspects of our inference strategy. First, we study the effect of relying on the instance branch if the segmentation decoder is unable to localize the target ($\max(s_s) < t_{s_s}$). Second, we study different values for $t_{s_c}$ that determines whether the target is detected by the instance model, see Tab.~\ref{tab:ablation_strategies}. 

If the segmentation branch cannot identify the target, using the instance branch improves tracking performance on all datasets (\eg~$+1.3\%$ on UAV123).
Furthermore, Tab.~\ref{tab:ablation_strategies} shows that our tracking pipeline achieves the best performance when setting $t_{s_c} = 0.3$ whereas smaller or larger values for $t_{s_c}$ decrease the tracking accuracy. Hence, it is important to find a suitable trade-off between frequently updating the model and memory to quickly adapt to appearance changes and updating only rarely to avoid contaminating the memory and model based on wrong predictions.

\subsection{Comparison to the state of the art}
\label{sec:sota_comparison}

Assessing segmentation accuracy on tracking datasets is not possible since only bounding box annotations are provided. Therefore, we compare our approach on six \gls{vot} benchmarks and validate the segmentation masks quality on two \gls{vos} datasets. 

\begin{figure}[t]
\centering
    \includegraphics[width=0.35\linewidth, keepaspectratio]{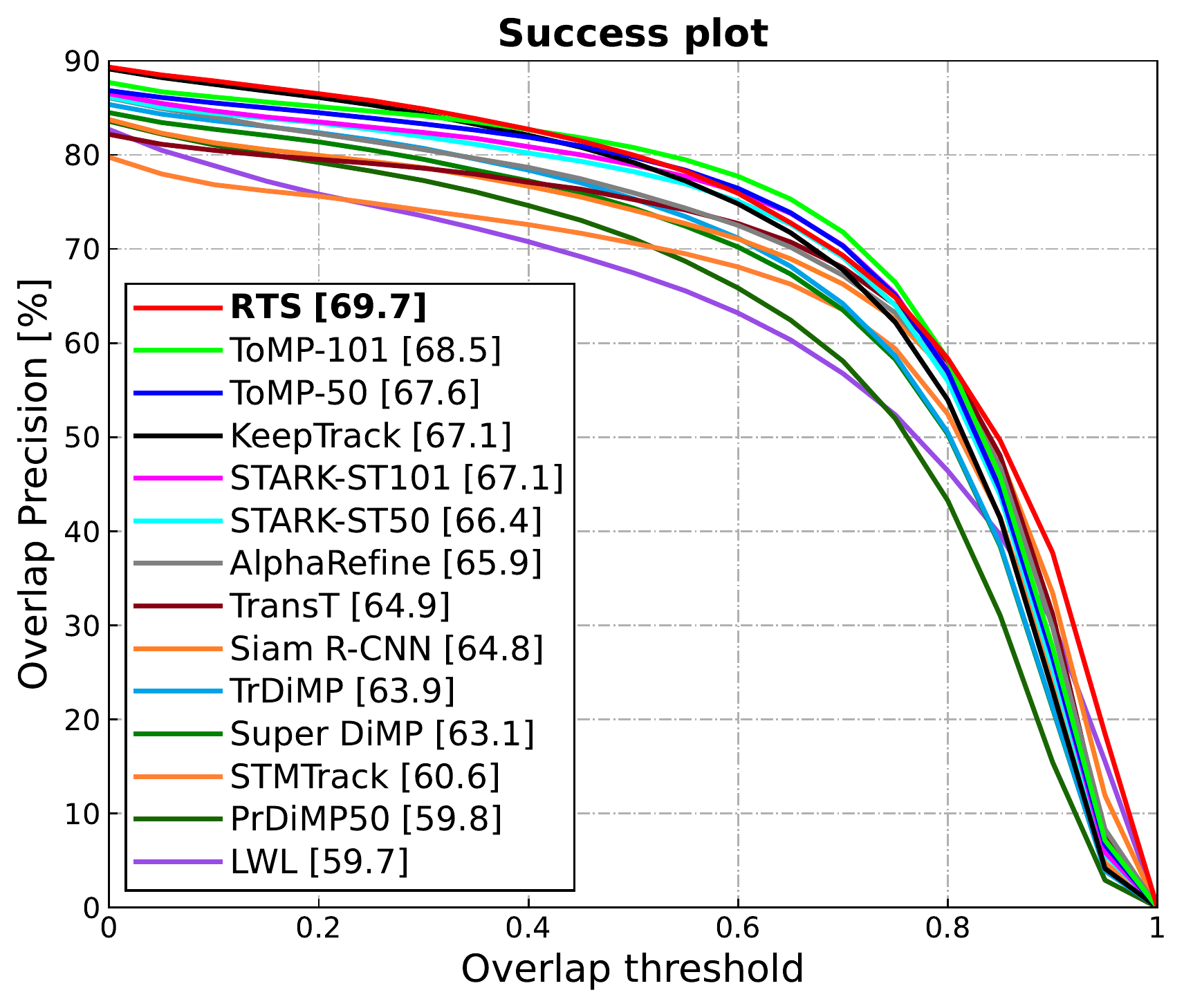}
    \includegraphics[width=0.35\linewidth, keepaspectratio]{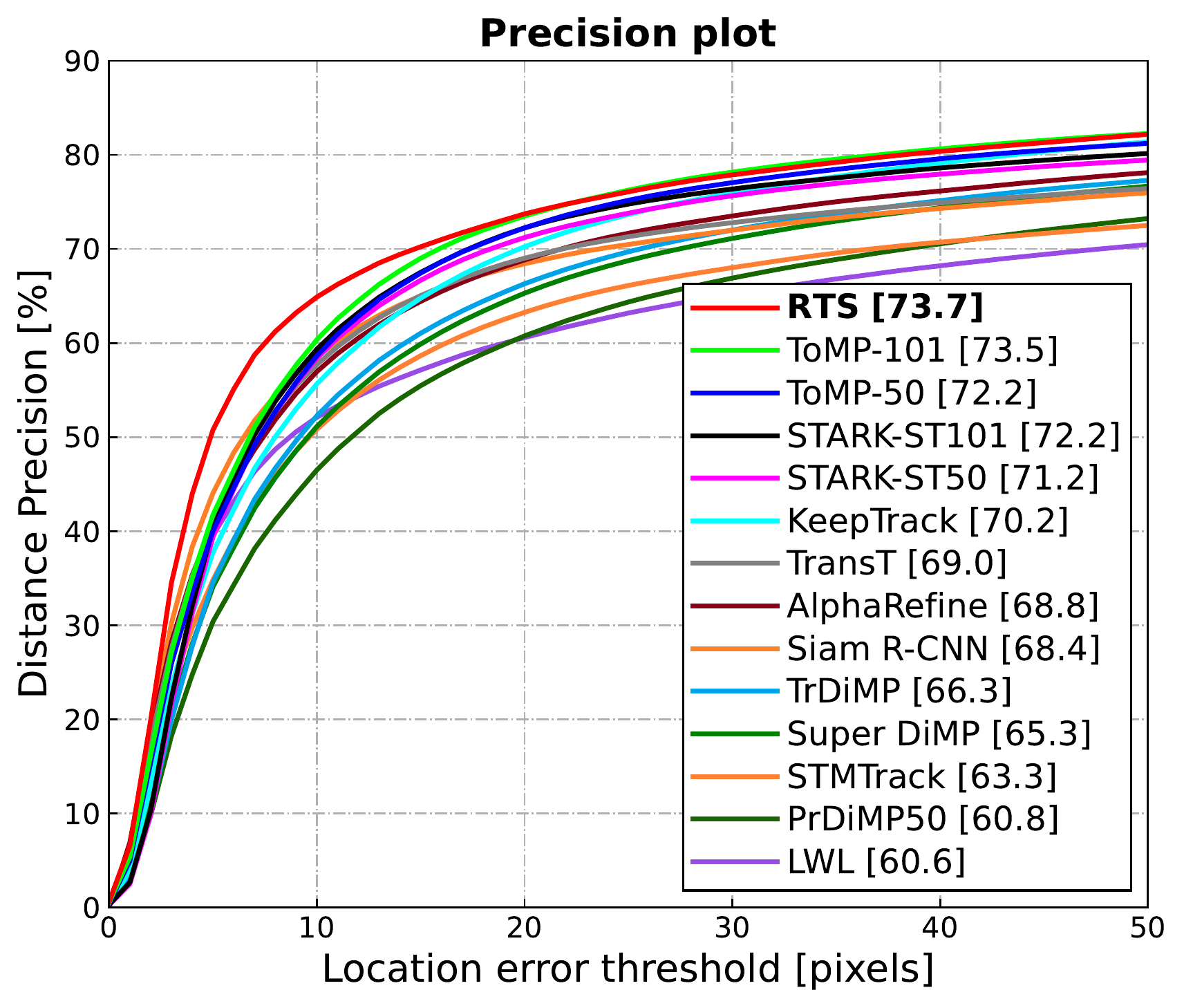}
\caption{Success (left) and Precision (right) plots on LaSOT~\cite{LaSOT} with other state-of-the-art methods. The AUCs for all methods are ordered and reported in the legend. Our method outperforms all existing approaches, both in Overlap Precision (left) and Distance Precision (right).}
\label{fig:success_lasot}
\end{figure}

\parsection{LaSOT~\cite{LaSOT}}
We evaluate our method on the test set of the LaSOT dataset, consisting of 280 sequences with 2500 frames on average. Thus, the benchmark challenges the long term adaptability and robustness of trackers. Fig.~\ref{fig:success_lasot} shows the success plot reporting the overlap precision $\text{OP}$ with respect to the overlap threshold $T$. Trackers are ranked by \gls{auc} score. In addition, Tab.~\ref{tab:lasot} reports the precision and normalized precision for all compared methods. Our method outperforms the state-of-the-art ToMP-50~\cite{Mayer_2022_CVPR_ToMP} and ToMP-101~\cite{Mayer_2022_CVPR_ToMP} by large margins ($+1.2\%$ and $+2.1\%$ \gls{auc} respectively). Our method is not only as robust as KeepTrack (see the success plot for $T<0.2$) but also estimates far more accurate bounding boxes than any tracker ($0.8<T<1.0$).

\begin{table}[!t]
	\caption{Comparison to the state of the art on the LaSOT~\cite{LaSOT} test set in terms of \gls{auc} score. The methods are ordered by \gls{auc} score.}
	\centering
	\newcommand{\best}[1]{\textbf{\textcolor{red}{#1}}}
	\newcommand{\scnd}[1]{\textbf{\textcolor{blue}{#1}}}
	\newcommand{\opt}[1]{\textbf{\textcolor{violet}{#1}}}
	\newcommand{\fast}[1]{\textbf{\textcolor{orange}{#1}}}

    \newcommand{\dist}{\enspace}
	\resizebox{1.00\columnwidth}{!}{%
        \begin{tabular}{l@{\dist}c@{\dist}c@{\dist}c@{\dist}c@{\dist}c@{\dist}c@{\dist}c@{\dist}c@{\dist}c@{\dist}c@{\dist}c@{\dist}c@{\dist}c@{\dist}c@{\dist}c@{\dist}c@{\dist}c@{\dist}c@{\dist}c@{\dist}c@{\dist}c@{\dist}c@{\dist}c@{\dist}c@{\dist}c@{\dist}c@{\dist}c@{\dist}c@{\dist}c@{\dist}c@{\dist}c@{\dist}c@{\dist}c@{\dist}c@{\dist}c@{\dist}c@{\dist}c@{\dist}c@{\dist}c@{\dist}c@{\dist}c@{\dist}c@{\dist}c@{\dist}c@{\dist}c@{\dist}c@{\dist}c@{\dist}c@{\dist}c@{\dist}c@{\dist}c@{\dist}c@{\dist}c@{\dist}c@{\dist}}
        	\toprule
        	               &               & ToMP & ToMP & Keep  & STARK  & Alpha  &        & Siam  & Tr   & Super & STM   & Pr   &     & DM    &      &      &       & \\
        	               & \textbf{RTS}  & 101  &  50 & Track & ST-101 & Refine & TransT & R-CNN & DiMP & DiMP  & Track & DiMP & LWL & Track & LTMU & DiMP & Ocean & D3S\\
        	               &               & \cite{Mayer_2022_CVPR_ToMP} & \cite{Mayer_2022_CVPR_ToMP} & \cite{mayer_ICCV_2021_keeptrack}& \cite{Yan_2021_ICCV_STARK} & \cite{Yan2021AlphaRefineBT} & \cite{Chen_2021_CVPR_TransT} & \cite{Voigtlaender_2020_CVPR_SiamRCNN} & \cite{Wang_2021_CVPR_TrDiMP} & \cite{pytracking} & \cite{Fu_2021_CVPR_STMTrack} & \cite{danelljan_ICCV_2019_PRDimP} & \cite{Bhat_2020_ECCV_LWL} & \cite{Zhang_2021_CVPR_DMTrack} & \cite{Dai_2020_CVPR_LTMU} & \cite{bhat_ICCV_2019_DiMP} & \cite{Zhang_2020_ECCV_Ocean} & \cite{Lukezic_CVPR_2020_D3S}  \\
        	\midrule
        	
        	Precision      & \best{73.7}  & \scnd{73.5} & 72.2        & 70.2 & 72.2 & 68.8 & 69.0 & 68.4 & 66.3 & 65.3 & 63.3 & 60.8 & 60.6 & 59.7 & 57.2 & 56.7 & 56.6 & 49.4 \\
        	Norm. Prec     &  76.2        & \best{79.2} & \scnd{78.0} & 77.2 & 76.9 & 73.8 & 73.8 & 72.2 & 73.0 & 72.2 & 69.3 & 68.8 & 63.3 & 66.9 & 66.2 & 65.0 & 65.1 & 53.9 \\
        	Success (AUC)  & \best{69.7}  & \scnd{68.5} & 67.6        & 67.1  & 67.1 & 65.9 & 64.9 & 64.8 & 63.9 & 63.1 & 60.6 & 59.8 & 59.7 & 58.4 & 57.2 & 56.9 & 56.0 & 49.2 \\
                	\midrule

        	$\Delta$ AUC to Ours   &   \textbf{-} & \up{1.2} & \up{2.1} &  \up{2.6}     & \up{2.6}         &\up{3.8}&\up{4.8}&\up{4.9}&\up{5.8}&\up{6.6}&\up{9.1}&\up{9.9}&\up{10.0} &\up{11.3}&\up{12.5}&\up{12.8}&\up{13.7}&\up{20.5}    \\
        	\bottomrule
        \end{tabular}
	}
	\label{tab:lasot}%
\end{table}
\begin{table}[!t]
	\caption{Results on the GOT-10k validation set  ~\cite{GOT10k}  in terms of Average Overlap (AO) and Success Rates (SR) for overlap thresholds of 0.5 and 0.75.}
	\centering
	\newcommand{\best}[1]{\textbf{\textcolor{red}{#1}}}
	\newcommand{\scnd}[1]{\textbf{\textcolor{blue}{#1}}}
	\newcommand{\dist}{\enspace}%
	\resizebox{0.6\columnwidth}{!}{%
        \begin{tabular}{l@{\dist}c@{\dist}c@{\dist}c@{\dist}c@{\dist}c@{\dist}c@{\dist}}
        	\toprule
        	          & \textbf{RTS}             &   STA       & \gls{lwl}  & PrDiMP-50         & DiMP-50            & SiamRPN++  \\
        	          &    &  \cite{zhao_ICCV_2021_vidboxseg_STA} &  \cite{Bhat_2020_ECCV_LWL} & \cite{danelljan_ICCV_2019_PRDimP} & \cite{bhat_ICCV_2019_DiMP}            & \cite{Li_2019_CVPR_SiamRPN++}    \\
        	\midrule
        	$\text{SR}_{0.50}(\%)$ &  \scnd{94.5} & \best{95.1}  & 92.4  & 89.6              & 88.7               & 82.8 \\
        	$\text{SR}_{0.75}(\%)$ &  \scnd{82.6} & \best{85.2}  & 82.2  & 72.8              & 68.8               &   -  \\
        	$\text{AO}(\%)$        &  \scnd{85.2} & \best{86.7}  & 84.6  & 77.8              & 75.3               & 73.0 \\\bottomrule
        \end{tabular}
	}
	\label{tab:got10k}%
\end{table}
\begin{table}[!t]
	\caption{Comparison to the state of the art on the TrackingNet~\cite{TrackingNet} test set in terms of AUC scores, Precision and Normalized Precision.}
	\centering
	\newcommand{\best}[1]{\textbf{\textcolor{red}{#1}}}
	\newcommand{\scnd}[1]{\textbf{\textcolor{blue}{#1}}}
	\newcommand{\dist}{\enspace}
	\resizebox{1.00\columnwidth}{!}{%
        \begin{tabular}{l@{\dist}c@{\dist}c@{\dist}c@{\dist}c@{\dist}c@{\dist}c@{\dist}c@{\dist}c@{\dist}c@{\dist}c@{\dist}c@{\dist}c@{\dist}c@{\dist}c@{\dist}c@{\dist}c@{\dist}c@{\dist}c@{\dist}c@{\dist}c@{\dist}c@{\dist}c@{\dist}c@{\dist}c@{\dist}c@{\dist}c@{\dist}c@{\dist}c@{\dist}c@{\dist}}
        	\toprule

        	           &                 & ToMP & ToMP & Keep     & STARK      & STARK &      &      &            & Siam        & Alpha  & STM   &     & Tr   & Super & Pr   &  \\
        	           & \textbf{RTS}    & 101  & 50   & Track    & ST101      & ST50  & STA & \gls{lwl} & TransT     & R-CNN       & Refine & Track & DTT & DiMP & DiMP  & DiMP & D3S \\
        	           &                 & \cite{Mayer_2022_CVPR_ToMP} & \cite{Mayer_2022_CVPR_ToMP} & \cite{mayer_ICCV_2021_keeptrack} & \cite{Yan_2021_ICCV_STARK} &\cite{Yan_2021_ICCV_STARK} & \cite{zhao_ICCV_2021_vidboxseg_STA} & \cite{Bhat_2020_ECCV_LWL} & \cite{Chen_2021_CVPR_TransT} &\cite{Voigtlaender_2020_CVPR_SiamRCNN} & \cite{Yan2021AlphaRefineBT} & \cite{Fu_2021_CVPR_STMTrack} &\cite{Yu_2021_ICCV_HPF} & \cite{Wang_2021_CVPR_TrDiMP} & \cite{pytracking} & \cite{danelljan_ICCV_2019_PRDimP} & \cite{Lukezic_CVPR_2020_D3S}  \\
        	\midrule
        	
        	Precision      & 79.4        & 78.9 & 78.6 & 73.8         & -           & -     & 79.1 & 78.4 & \best{80.3} & \scnd{80.0} & 78.3   & 76.7  & 78.9 & 73.1 & 73.3 & 70.4 & 66.4  \\
        	Norm. Prec     & 86.0        & 86.4 & 86.2 & 83.5         & \best{86.9} & 86.1  & 84.7 & 84.4 & \scnd{86.7} & 85.4        & 85.6   & 85.1  & 85.0 & 83.3 & 83.5 & 81.6 & 76.8  \\
        	Success (AUC)  & \scnd{81.6} & 81.5 & 81.2 & 78.1         & \best{82.0} & 81.3  & 81.2 & 80.7 & 81.4        & 81.2        & 80.5   & 80.3  & 79.6 & 78.4 & 78.1 & 75.8 & 72.8  \\
        	\midrule
        	$\Delta$ AUC to Ours & \textbf{-} & \up{0.1} & \up{0.4} & \up{3.5}& \down{0.4} &\up{0.3}&\up{0.4}&\up{0.9}&\up{0.2}&\up{0.4}&\up{1.1}&\up{1.3}&\up{2.0}&\up{3.2}&\up{3.5}&\up{5.8}&\up{8.8}    \\

        	\bottomrule
        \end{tabular}
	}
	\label{tab:trackingnet}%
\end{table}
\begin{table}[t]
	\caption{Comparison with state-of-the-art on the UAV123~\cite{Mueller_2016_ECCV_UAV123} and NFS~\cite{Galoogahi_2017_ICCV_NFS} datasets in terms of AUC score.}
	\centering
	\newcommand{\best}[1]{\textbf{\textcolor{red}{#1}}}
	\newcommand{\scnd}[1]{\textbf{\textcolor{blue}{#1}}}
	\newcommand{\opt}[1]{\textbf{\textcolor{violet}{#1}}}
	\newcommand{\fast}[1]{\textbf{\textcolor{orange}{#1}}}
	\newcommand{\dist}{\enspace}
	\resizebox{1.00\columnwidth}{!}{%
        \begin{tabular}{l@{\dist}c@{\dist}c@{\dist}c@{\dist}c@{\dist}c@{\dist}c@{\dist}c@{\dist}c@{\dist}c@{\dist}c@{\dist}c@{\dist}c@{\dist}c@{\dist}c@{\dist}c@{\dist}c@{\dist}c@{\dist}c@{\dist}c@{\dist}c@{\dist}c@{\dist}c@{\dist}c@{\dist}c@{\dist}c@{\dist}c@{\dist}c@{\dist}c@{\dist}c@{\dist}c@{\dist}c@{\dist}c@{\dist}}
        	\toprule
        	        &  & ToMP & ToMP & Keep        &             & STARK &        &            & STARK     & Super  & Pr    & STM        & Siam & Siam  &      &      & \\
        	        &    \textbf{RTS}          & 101  & 50   & Track       & CRACT       & ST101 & TrDiMP & TransT     & ST50      & DiMP   & DiMP  & Track      & AttN & R-CNN & KYS  & DiMP & LWL \\
        	        &              & \cite{Mayer_2022_CVPR_ToMP} & \cite{Mayer_2022_CVPR_ToMP} & \cite{mayer_ICCV_2021_keeptrack} & \cite{Fan_2020_arxiv_CRACT} & \cite{Yan_2021_ICCV_STARK} & \cite{Wang_2021_CVPR_TrDiMP} & \cite{Chen_2021_CVPR_TransT} & \cite{Yan_2021_ICCV_STARK} & \cite{pytracking} & \cite{danelljan_ICCV_2019_PRDimP} & \cite{Fu_2021_CVPR_STMTrack} & \cite{Yu_2020_CVPR_SiamAttN} & \cite{Voigtlaender_2020_CVPR_SiamRCNN} & \cite{Bhat_2020_ECCV_KYS} & \cite{bhat_ICCV_2019_DiMP} & \cite{Bhat_2020_ECCV_LWL}\\          
        	\midrule
        	UAV123  & 67.6 & 66.9        & 69.0        & \best{69.7} & 66.4        & 68.2  & 67.5   & \scnd{69.1}& \scnd{69.1}& 67.7   & 68.0 & 64.7        & 65.0 & 64.9  & --   & 65.3 & 59.7\\
        	NFS     & 65.4 & \scnd{66.7} & \best{66.9} & 66.4        & 62.5        & 66.2  & 66.2   & 65.7       & 65.2       & 64.8   & 63.5 & --          & --   & 63.9  & 63.5 & 62.0 & 61.5\\
            \bottomrule
        \end{tabular}
	}
	\label{tab:uav_nfs}
\end{table}

\parsection{GOT-10k~\cite{GOT10k}}
The large-scale GOT-10k dataset contains over 10.000 shorter sequences. Since we train our method on several datasets instead of only GOT-10k \emph{train}, we evaluate it on the \emph{val} set only, which consists of 180 short videos. We compile the results in Tab.~\ref{tab:got10k}. Our method ranks second for all metrics, falling between two VOS-oriented methods, $+0.6\%$ over \gls{lwl}~\cite{Bhat_2020_ECCV_LWL} and $-1.5\%$ behind STA~\cite{zhao_ICCV_2021_vidboxseg_STA}. Our tracker outperforms other trackers by a large margin.

\parsection{TrackingNet~\cite{TrackingNet}}
We compare our approach on the test set of the TrackingNet dataset, consisting of 511 sequences. Tab.~\ref{tab:trackingnet} shows the results obtained from the online evaluation server. Our method outperforms most of the existing approaches and ranks second in terms of \gls{auc}, close behind STARK-ST101~\cite{Yan_2021_ICCV_STARK} which is based on a ResNet-101 backbone. Note that we outperform STARK-ST50~\cite{Yan_2021_ICCV_STARK} that uses a ResNet-50 as backbone. Also, we achieve a higher precision score than other methods that produce a segmentation mask output such as \gls{lwl}~\cite{Bhat_2020_ECCV_LWL}, STA~\cite{zhao_ICCV_2021_vidboxseg_STA}, Alpha-Refine~\cite{Yan2021AlphaRefineBT} and D3S~\cite{Lukezic_CVPR_2020_D3S}.

\parsection{UAV123~\cite{Mueller_2016_ECCV_UAV123}}
The UAV dataset consists of 123 test videos that contain small objects, target occlusion, and distractors. Small objects are particularly challenging in a segmentation setup. Tab.~\ref{tab:uav_nfs} shows the achieved results in terms of success \gls{auc}. Our method achieves competitive results on UAV123, close to TrDiMP~\cite{Wang_2021_CVPR_TrDiMP} or SuperDiMP~\cite{pytracking}. It outperforms LWL~\cite{Bhat_2020_ECCV_LWL} by a large margin.

\parsection{NFS~\cite{Galoogahi_2017_ICCV_NFS}}
The NFS dataset (30FPS version) contains 100 test videos with fast motions and challenging sequences with distractors. Our method achieves an \gls{auc} score that is only 1\% below the current best method KeepTrack~\cite{mayer_ICCV_2021_keeptrack} while outperforming numerous other trackers, including STARK-ST50~\cite{Yan_2021_ICCV_STARK} (+0.2) SuperDiMP~\cite{bhat_ICCV_2019_DiMP} (+0.6) and PrDiMP~\cite{danelljan_ICCV_2019_PRDimP} (+1.9).

\parsection{VOT 2020~\cite{Kristan_2020_ECCVW_VOT2020}}
Finally, we evaluate our method on the VOT2020 short-term challenge. It consists of 60 videos and provides segmentation mask annotations. For the challenge, the multi-start protocol is used and the tracking performance is assessed based on accuracy and robustness. We compare with the top methods on the leader board and include more recent methods in Tab.~\ref{tab:vot2020st}. In this setup, our method ranks \nth{2} in Robustness, thus outperforming most of the other methods. In particular, we achieve a higher EAO score than STARK~\cite{Yan_2021_ICCV_STARK}, LWL~\cite{Bhat_2020_ECCV_LWL}, AlphaRefine~\cite{Yan2021AlphaRefineBT} and D3S~\cite{Lukezic_CVPR_2020_D3S}.

\begin{table}[!t]
	\caption{Results on the VOT2020-ST ~\cite{Kristan_2020_ECCVW_VOT2020} challenge in terms of Expected Average Overlap (EAO), Accuracy and Robustness.}
	\centering
	\newcommand{\best}[1]{\textbf{\textcolor{red}{#1}}}
	\newcommand{\scnd}[1]{\textbf{\textcolor{blue}{#1}}}
	\newcommand{\dist}{\enspace}
	\resizebox{1.00\columnwidth}{!}{%
        \begin{tabular}{l@{\dist}c@{\dist}c@{\dist}c@{\dist}c@{\dist}c@{\dist}c@{\dist}c@{\dist}c@{\dist}c@{\dist}c@{\dist}c@{\dist}c@{\dist}}
        	\toprule
        	          &                  & STARK        & STARK-       &           &              &              &        &          &               &       &       &        \\
        	          &                  & ST-50        & ST-101-      &           &              & Ocean        & Fast   & Alpha    &               &       &       &        \\
        	          & \textbf{RTS}     & +AR          & +AR          & \gls{lwl} & STA         & Plus         & Ocean  & Refine   & RPT           & AFOD  & D3S   & STM    \\
        	          &                  &  \cite{Yan_2021_ICCV_STARK} & \cite{Yan_2021_ICCV_STARK}   & \cite{Kristan_2020_ECCVW_VOT2020} & \cite{zhao_ICCV_2021_vidboxseg_STA} & \cite{Kristan_2020_ECCVW_VOT2020} & \cite{Kristan_2020_ECCVW_VOT2020} & \cite{Kristan_2020_ECCVW_VOT2020} & \cite{Kristan_2020_ECCVW_VOT2020} & \cite{Kristan_2020_ECCVW_VOT2020} & \cite{Kristan_2020_ECCVW_VOT2020} & \cite{Kristan_2020_ECCVW_VOT2020} \\
        	\midrule
        	Robustness   &  \scnd{0.845}   & 0.817        & 0.789        & 0.798   & 0.824        & 0.842 & 0.803 &  0.777   & \best{ 0.869} & 0.795 & 0.769 & 0.574 \\
        	Accuracy     &  0.710   & \scnd{0.759} & \best{0.763} & 0.719   & 0.732        & 0.685        & 0.693 &  0.754   & 0.700         & 0.713 & 0.699 & 0.751 \\
        	EAO          &  0.506   & 0.505        & 0.497        & 0.463   & \scnd{0.510} & 0.491        & 0.461 &  0.482   & \best{0.530}  & 0.472 & 0.439 & 0.308 \\
\midrule
    $\Delta$ EAO to Ours & \textbf{-}    & \up{0.001} & \up{0.009}	 & \up{0.043}	 & \down{0.004}	 & \up{0.015}	 & \up{0.045}	 & \up{0.024}	 & \down{0.024}	 & \up{0.034}	 & \up{0.067}	 & \up{0.198}    \\

        	\bottomrule
        \end{tabular}
	}
	\label{tab:vot2020st}%
\end{table}

\begin{table}[!t]
	\caption{Results on the Youtube-VOS 2019~\cite{xu2018youtubevos} and DAVIS 2017~\cite{Pont-Tuset_arXiv_2017_DAVIS} datasets. The table is split in two parts to separate methods using bounding box initialization or segmentation masks initialization, in order to enable a fair comparison.}
	\centering
	\newcommand{\best}[1]{\textbf{\textcolor{red}{#1}}}
	\newcommand{\scnd}[1]{\textbf{\textcolor{blue}{#1}}}
    \newcommand{\dist}{\qquad}%
	\resizebox{1.00\columnwidth}{!}{%
        
    \begin{tabular}{l@{\dist}c@{\dist}c@{\dist}c@{\dist}c@{\dist}c|c@{\dist}c@{\dist}c@{\quad}}
    % \begin{tabular}{l@{~~~}c@{~~}c@{~~}c@{~~}c@{~~}c|c@{~~}c@{~~}c}
    \toprule
    \multirow{2}{*}{Method} & \multicolumn{5}{c}{\textbf{YouTube-VOS 2019~\cite{xu2018youtubevos}}} & \multicolumn{3}{c}{\textbf{DAVIS 2017~\cite{Pont-Tuset_arXiv_2017_DAVIS}}} \\
     &  $\mcG$  & $\mcJ_\text{seen}$  & $\mcJ_\text{unseen}$ & $\mcF_\text{seen}$ & ~$\mcF_\text{unseen}$~~ & ~~$\mathcal{J \& F}$~ & $\mathcal{J}$  & $\mathcal{F}$ \\
    \midrule
    {\bf RTS}                                       &   79.7      & 77.9       & 75.4         &       82.0        & 83.3  &    80.2     &  77.9       &    82.6  \\
    \gls{lwl}~\cite{Bhat_2020_ECCV_LWL}              & 81.0        & 79.6       &        76.4  &       83.8  &       84.2  &       81.6  &      79.1   &     84.1 \\
    STA~\cite{zhao_ICCV_2021_vidboxseg_STA}          & 80.6        & -          &         -    &       -     &       -     &       -     &      -      &     -    \\
    STM~\cite{Oh_2019_ICCV_STM}                      & 79.2        & 79.6       &	     73.0  &	   83.6	 &       80.6  &       81.8  &      79.2   &     84.3 \\
    \midrule
    \midrule
    {\bf RTS (Box)}                                 & 70.8        & 71.1       &        65.2  &       74.0  &       72.8  &       72.6  &      69.4   &     75.8 \\
    \gls{lwl} (Box)~\cite{Bhat_2020_ECCV_LWL}        & -           & -          &        -     &       -     &       -     &       70.6  &      67.9   &     73.3 \\
    Siam-RCNN~\cite{Voigtlaender_2020_CVPR_SiamRCNN} & 67.3        & 68.1       &        61.5  &       70.8  &       68.8  &       70.6  &      66.1   &     75.0 \\
    D3S~\cite{wang2019fast_SiamMask}                 & -           & -          &	     -     &	   -	 &       -     &       60.8  &      57.8   &     63.8 \\
    SiamMask~\cite{Lukezic_CVPR_2020_D3S}            & 52.8        & 60.2       &	     45.1  &	   58.2	 &       47.7  &       56.4  &      54.3   &     58.5 \\

    \bottomrule
    \end{tabular}       
	}
	\label{tab:ytvos_davis}%
\end{table}

\parsection{YouTube-VOS 2019~\cite{xu2018youtubevos}}
We use the validation set which consist of 507 sequences. They contain 91 object categories out of which 26 are \emph{unseen} in the training set. The results presented in Tab.~\ref{tab:ytvos_davis} were generated by an online server after uploading the raw results.
On this benchmark, we want to validate the quality of the produced segmentation masks rather than to achieve the best accuracy possible. Hence, we use the same model weight as for VOT without further fine tuning. 

When using the provided segmentation masks for initialization, we observe that our method performs slightly worse than LWL~\cite{Bhat_2020_ECCV_LWL} and STA~\cite{zhao_ICCV_2021_vidboxseg_STA} (-1.3 $\mcG$, -0.9 $\mcG$) but still outperforms the VOS method STM~\cite{Oh_2019_ICCV_STM} (+0.5 $\mcG$). We conclude that our method can generate accurate segmentation masks. When using bounding boxes to predict both the initialization and segmentation masks, we outperform all other methods by a large margin. This confirms that even with our bounding-box initialization strategy, RTS produces accurate segmentation masks.

\parsection{DAVIS 2017~\cite{Pont-Tuset_arXiv_2017_DAVIS}}
Similarly, we compare our method on the validation set of DAVIS 2017~\cite{Pont-Tuset_arXiv_2017_DAVIS}, which contains 30 sequences. We do not fine tune the model for this benchmark. The results are shown in Tab.~\ref{tab:ytvos_davis} and confirm the observation made above that RTS is able to generate accurate segmentation masks. Our method is competitive in the mask-initialization setup. In the box-initialization setup however, our approach outperforms all other methods in $\mathcal{J \& F}$, in particular the segmentation trackers like SiamMask~\cite{wang2019fast_SiamMask} (+16.2) and  D3S~\cite{Lukezic_CVPR_2020_D3S} (+11.8).

%-------------------------------------------------------------------------
\section{Conclusion}

We introduced RTS, a robust, end-to-end trainable, segmentation-driven tracking method that is able to generate accurate segmentation masks. Compared to the traditional bounding box outputs of classical visual object trackers, segmentation masks enable a more accurate representation of the target's shape and extent. The proposed instance localization branch helps increasing the robustness of our tracker to enable reliable tracking even for long sequences of thousands of frames.
Our method outperforms previous segmentation-driven tracking methods by a large margin, and it is competitive on several VOT benchmarks. In particular, we set a new state of the art on the challenging LaSOT~\cite{LaSOT} dataset with a success AUC of 69.7\%. Competitive results on two VOS datasets confirm the high quality of the generated segmentation masks.\\

\parsection{Acknowledgements} This work was partly supported by uniqFEED AG and the ETH Future Computing Laboratory (EFCL) financed by a gift from Huawei Technologies.

\clearpage
\bibliographystyle{splncs04}
\bibliography{egbib}

\clearpage

\section*{Appendix}
\beginappendix
In this Appendix, we provide further details on various aspects of our tracking pipeline. First, we provide additional architectural and inference details in Sections~\ref{sup:sec:architecture} and \ref{sup:sec:inference}. Second, we provide additional ablation studies, in particular on the loss weighting parameter $\eta$ on different benchmarks to show the importance of the auxiliary instance localization loss in Section~\ref{sup:sec:ablation}. Then, we provide success plots for different \gls{vot} benchmarks as well as a detailed analysis of our results on LaSOT~\cite{LaSOT} by comparing our approach against the other state-of-the-art methods for all the dataset attributes in Section~\ref{sup:sec:attributes}. Finally, we provide some additional visual comparison to other trackers in Section~\ref{sup:sec:morecontent}.

\section{Additional Architecture details}
\label{sup:sec:architecture}

\begin{figure}[t]
\centering
\includegraphics[width=1.0\linewidth]{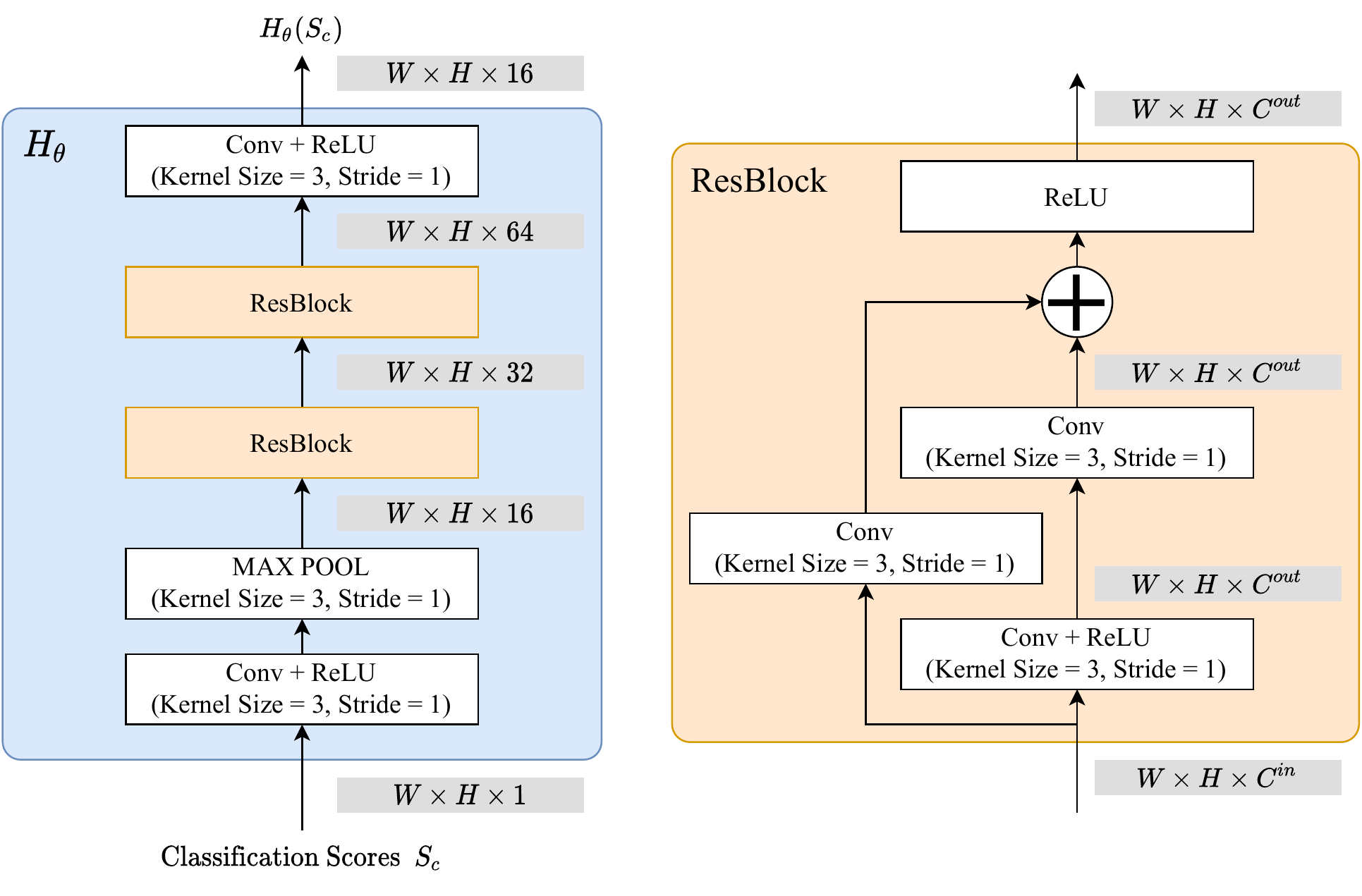}
\caption{Classification Scores Encoder $\clfScoresEnc$.}%
\label{sup:fig:scores_encoder}
\end{figure}

\parsection{Classification Scores Encoder $\clfScoresEnc$} First, we describe in Figure~\ref{sup:fig:scores_encoder} the architecture of the Classification Scores Encoder $\clfScoresEnc$. It takes as input the $H\times W$-dimensional scores predicted by the Instance Localization (\emph{Classification}) branch and outputs a 16 channels deep representation of those scores. The score encoder consists of a convolutional layer followed by a max-pool layer with stride one and two residual blocks. The output of the residual blocks has 64 channels. Thus, the final convolutional layer reduces the number of channels of the output to 16 to match the encoded scores with the mask encoding. All the convolutional layers use ($3\times3$) kernels with a stride of one to preserve the spatial size of the input classification scores.

\parsection{Segmentation Decoder $\segDecoder$} The segmentation decoder has the same structure has in LWL~\cite{Bhat_2020_ECCV_LWL}. Together with the backbone, it shows a U-Net structure and mainly consists of four decoder blocks. It takes as input the extracted ResNet-50 backbone features and the combined encoding $x_\text{f}$ from both the instance localization branch ($\clfScoresEnc(\clfscores)$) and the segmentation branch ($\maskencoding$), with $x_\text{f} = \maskencoding + \clfScoresEnc(\clfscores)$. Since the encoded instance localization scores have a lower spatial resolution than the mask encoding $x_\text{m}$, we upscale the encoded instance localization scores using a bilinear interpolation before adding it with the mask encoding $x_\text{m}$. We refer the reader to \cite{Bhat_2020_ECCV_LWL} for more details about the decoder structure.

\parsection{Segmentation Branch} We use the same architectures for the feature extractor $F_\theta$, the label encoder $E_\theta$, the weight predictor $W_\theta$, the few-shot learner $A_\theta$ and the segmentation model $T_\tau$ as proposed in LWL~\cite{Bhat_2020_ECCV_LWL}. Hence, we refer the reader to~\cite{Bhat_2020_ECCV_LWL} for more details.

\parsection{Instance Localization Branch} We use the same architectures for the feature extractor $G_\theta$, the model predictor $P_\theta$ and the instance model $T_\kappa$ as proposed in DiMP~\cite{bhat_ICCV_2019_DiMP}. Hence, we refer the reader to~\cite{bhat_ICCV_2019_DiMP} for more details. 

\section{Additional Inference details}
\label{sup:sec:inference}

\parsection{Search region selection} The backbone does not extract features on the full image. Instead, we sample a smaller image patch for extraction, which is centered at the current target location and 6 times larger than the current estimated target size, when it does not exceed the size of the image. The estimation of the target state (position and size) is therefore crucial to ensure an optimal crop. In most situations, the segmentation output is used to determine the target state since it has a high accuracy. The \emph{target center} is computed as the center of mass of the predicted per-pixel segmentation probability scores. The \emph{target size} is computed as the variance of the segmentation probability scores.
    
If the segmentation branch cannot find the target (as described in the main paper), but the instance branch still outputs a high enough confidence score, we use it to update the target position. This is particularly important in sequences where the target is becoming too small for some time, but we can still track the target position.
    
When both branch cannot find the target, the internal state of the tracker is not updated. We upscale the search area based on the previous 60 valid predicted scales. This is helpful in situations where the size of the object shrinks although its size does not change. This typically happens during occlusions, or if the target goes out of the frame partially or completely.

\section{Additional Ablations}
\label{sup:sec:ablation}

In this section, we provide additional ablation studies related to our method, first on the weighting of the segmentation and classification losses used for training, second on the parameters that might make a difference specifically for \gls{vos} benchmarks like Youtube-VOS~\cite{xu2018youtubevos}.

\parsection{Weighting segmentation and classification losses}
For this ablation, we study the weighting of the segmentation loss $\mathcal{L}_\text{s}$ and the instance localization loss $\mathcal{L}_\text{c}$ in the total loss $\mathcal{L}_\text{tot}$. It used to train our model and its influence on the overall performance during tracking. We recall that

\begin{equation}
    \mathcal{L}_\text{tot} =  \mathcal{L}_\text{s} + \eta \cdot\mathcal{L}_\text{c}.
\label{sup:eq:train_combined_loss}
\end{equation}
Table~\ref{sup:tab:ablation_weights} shows the results when training the tracker with three different values of $\eta$ on five \gls{vot} datasets. First, we examine the case where we omit the auxiliary instance localization loss ($\eta=0.0$). This means that the whole pipeline is trained for segmentation and the instance branch is not trained to produce specifically accurate localization scores. We observe that this setting leads to the lowest performance on all tested datasets, often by a large margin.
Secondly, we test a dominant segmentation loss ($\eta=0.4$), since the segmentation branch needs to be trained for a more complex task than the instance branch. We see a performance gain for almost all datasets. Thus, employing the auxiliary loss to train the instance localization branch helps to improve the tracking performance. We observe that using the auxiliary loss leads to localization scores generated during inference that are sharper, cleaner and localize the center of the target more accurately.
Finally, we put an even higher weight on the classification term ($\eta=10$). This setup leads to an even more accurate localization, and leads to the best results on average. Thus, we set $\eta = 10$ to train our tracking pipeline.

\begin{table}[t]
    \caption{Ablation on the classification vs. segmentation loss weighting on different datasets in terms of AUC (area-under-the-curve) and AO (average overlap)}
    \centering
    \newcommand{\dist}{\qquad}
    \resizebox{1.0\columnwidth}{!}{%
        \begin{tabular}{c@{\dist}|c@{\dist}c@{\dist}c@{\dist}c@{\dist}c@{\dist}c@{\dist}}
        \toprule
        
                    & LaSOT~\cite{LaSOT} & GOT-10k~\cite{GOT10k} & TrackingNet~\cite{TrackingNet} & NFS~\cite{Galoogahi_2017_ICCV_NFS} & UAV123~\cite{Mueller_2016_ECCV_UAV123} \\
         $\eta$     & AUC                & AO                    &  AUC                           & AUC                                & AUC   \\ 
        \midrule
          0.0       &  67.7              &   84.0                & 81.2                             & 63.7                               & 64.7  \\
          0.4       &  69.8              &   84.0                & 81.4                             & 66.2                               & 67.4  \\ 
          10        &  69.7              &   85.2                & 81.6                             & 65.4                               & 67.6   \\ \bottomrule

    \end{tabular}}
    \label{sup:tab:ablation_weights}
\end{table}

\begin{table}[!t]
	\caption{Results on the Youtube-VOS 2019~\cite{xu2018youtubevos} and DAVIS 2017~\cite{Pont-Tuset_arXiv_2017_DAVIS} datasets with a fined tuned model and inference parameters refered as \emph{RTS (YT-FT)}.}
	\centering
	\newcommand{\best}[1]{\textbf{\textcolor{red}{#1}}}
	\newcommand{\scnd}[1]{\textbf{\textcolor{blue}{#1}}}
	\newcommand{\dist}{\qquad}
	\resizebox{1.00\columnwidth}{!}{%
        
    \begin{tabular}{l@{\dist}c@{\dist}c@{\dist}c@{\dist}c@{\dist}c|c@{\dist}c@{\dist}c}
    \toprule
    & \multicolumn{5}{c}{\textbf{YouTube-VOS 2019~\cite{xu2018youtubevos}}} & \multicolumn{3}{c}{\textbf{DAVIS 2017~\cite{Pont-Tuset_arXiv_2017_DAVIS}}} \\
    Method &  $\mcG$  & $\mcJ_\text{seen}$  & $\mcJ_\text{unseen}$ & $\mcF_\text{seen}$ & ~$\mcF_\text{unseen}$~~ & ~~$\mathcal{J \& F}$~ & $\mathcal{J}$  & $\mathcal{F}$ \\
    \midrule
    {\bf RTS}                                       & 79.7        & 77.9       &        75.4  &       82.0  &       83.3  &       80.2  &      77.9   & 82.6 \\
    {\bf RTS (YT-FT)}                           & 80.3        & 78.8       &        76.2  &       82.9  &       83.5  &       80.3  &      77.7   & 82.9 \\
    LWL~\cite{Bhat_2020_ECCV_LWL}              & 81.0        & 79.6       &        76.4  &       83.8  &       84.2  &       81.6  &      79.1   & 84.1 \\
    STA~\cite{zhao_ICCV_2021_vidboxseg_STA}          & 80.6        & -          &         -    &       -     &       -     &       -     &      -      &  -   \\
    STM~\cite{Oh_2019_ICCV_STM}                      & 79.2        & 79.6       &	     73.0  &	   83.6	 &       80.6  &       81.8  &      79.2   & 84.3 \\

    \bottomrule
    \end{tabular}       
	}
	\label{sup:tab:ytvos_davis_ft}%
\end{table}

\parsection{Fine-tuning on Youtube-VOS~\cite{LaSOT}}
\label{sup:sec:finetuned}
In this section, we analyze whether we can gear our pipeline towards \gls{vos} benchmarks. To do that, we take our model and inference parameters, and modify them slightly. On the one hand, the model is fined-tuned for 50 epochs using Youtube-VOS~\cite{xu2018youtubevos} only for both training and validation. We also increase the initialization phase from 100 to 200 frames, and remove the relative target scale change limit from one frame to the next. In our standard model, we limit that scale change to 20\% for increased robustness.

The results are presented in Table~\ref{sup:tab:ytvos_davis_ft} for Youtube-VOS~\cite{xu2018youtubevos} and Davis~\cite{Pont-Tuset_arXiv_2017_DAVIS}. We observe that the performances between both of our models stay very close for Davis, but that the fine-tuned model is getting closer to the baseline LWL~\cite{Bhat_2020_ECCV_LWL} for Youtube-VOS. The more frequent updates seem to help, and not restricting the scale change of objects from frame to frames seems to play a role, since we get an improvement of 0.6 in $\mcG$ score.

\begin{figure}[t]
\centering
    \includegraphics[width=0.45\linewidth, keepaspectratio]{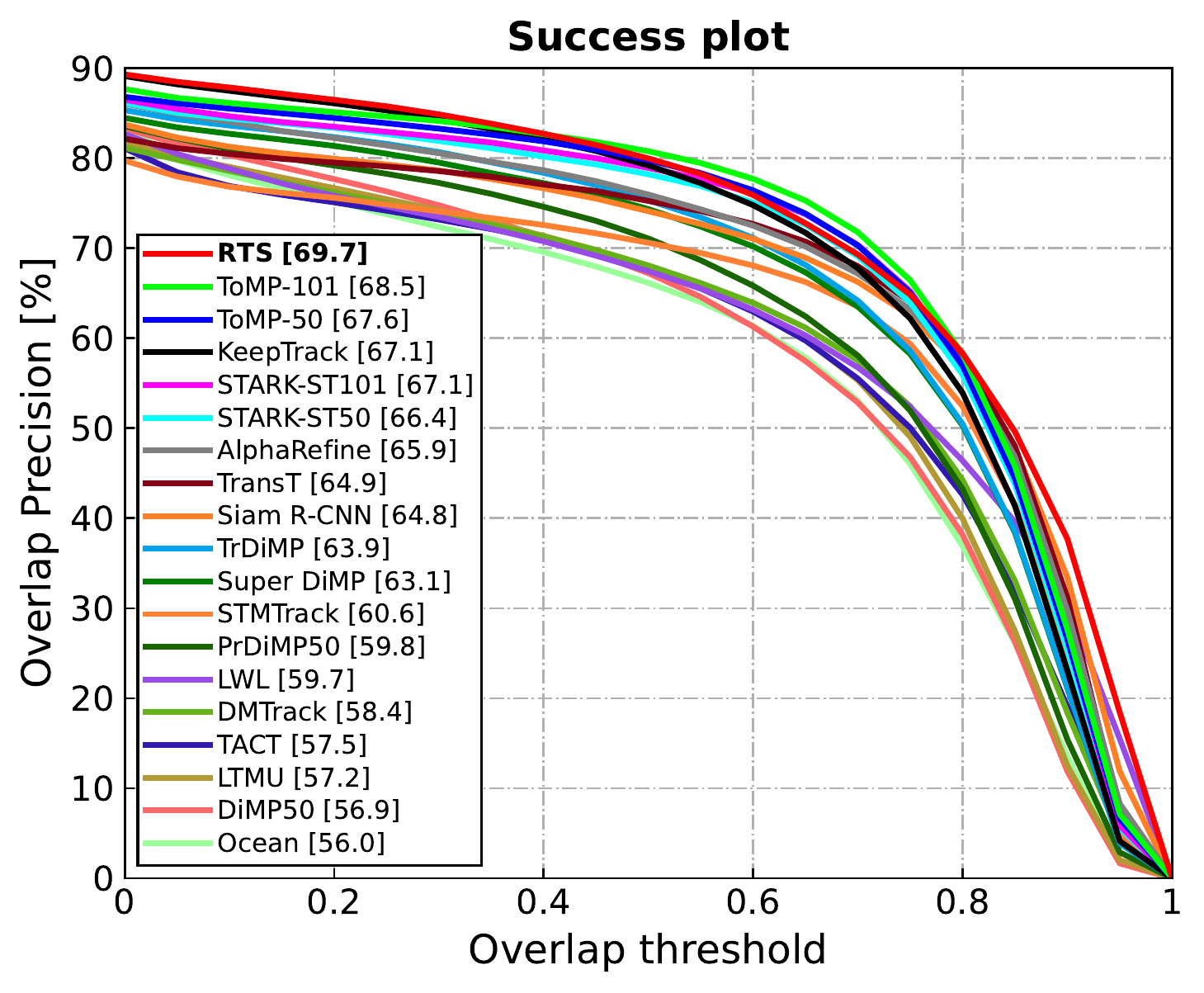}
    \includegraphics[width=0.45\linewidth, keepaspectratio]{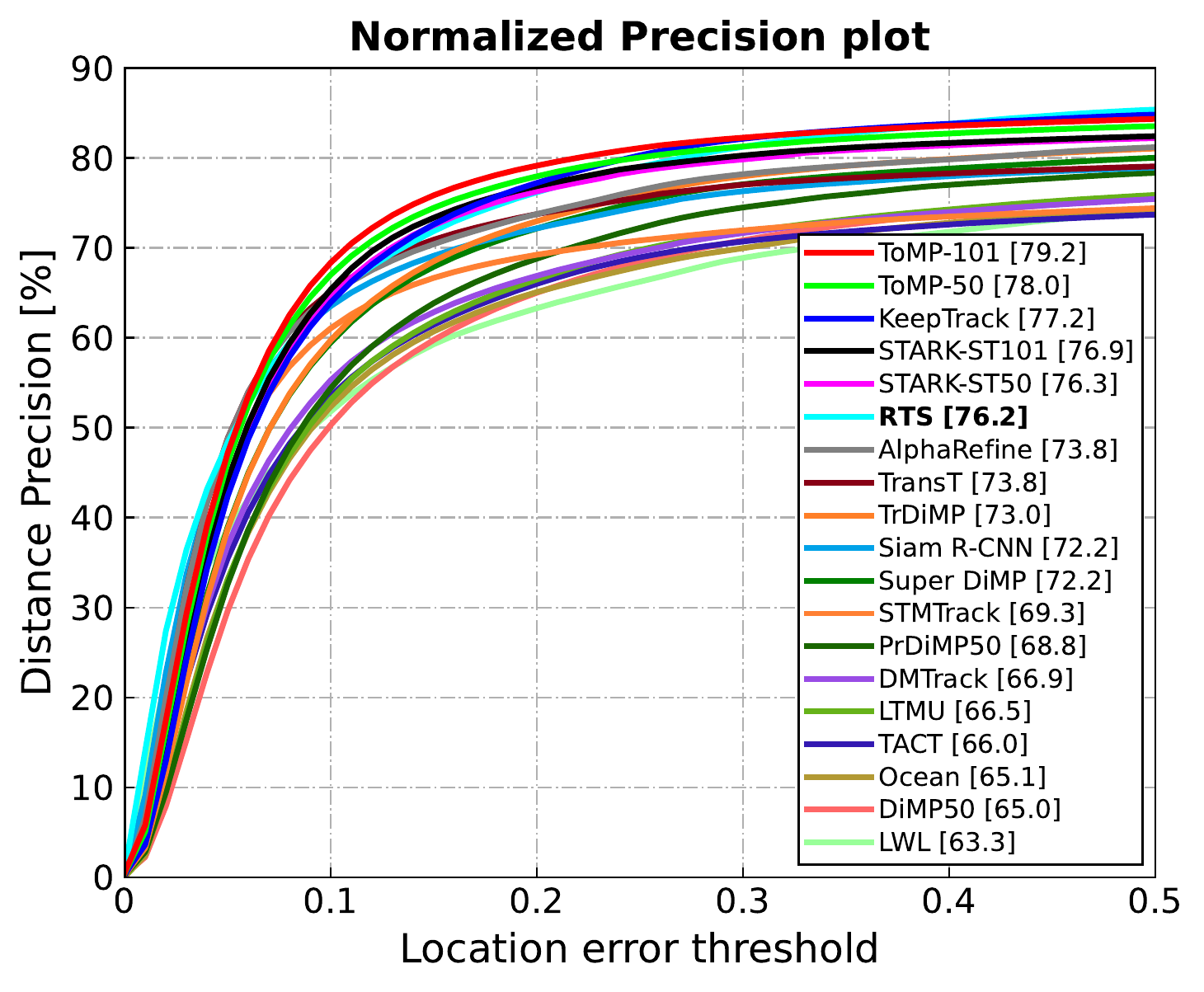}
    \includegraphics[width=0.45\linewidth, keepaspectratio]{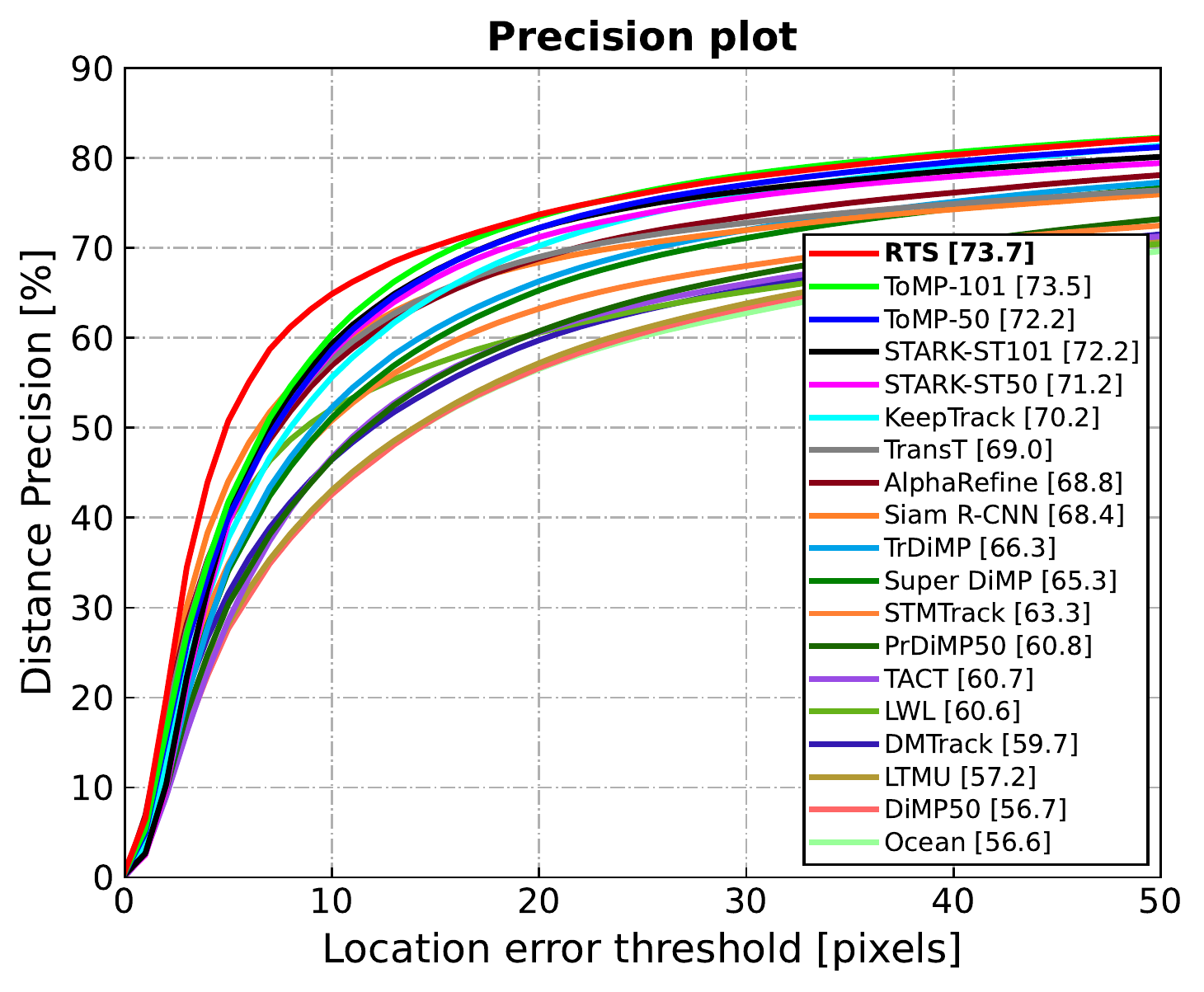}
\caption{Success, precision and normalized precision plots on LaSOT~\cite{LaSOT}. Our approach outperforms all other methods by a large margin in AUC, reported in the legend.}
\label{sup:fig:lasot_plots}
\end{figure}
\begin{figure}[t]
\centering
    \includegraphics[width=0.45\linewidth, keepaspectratio]{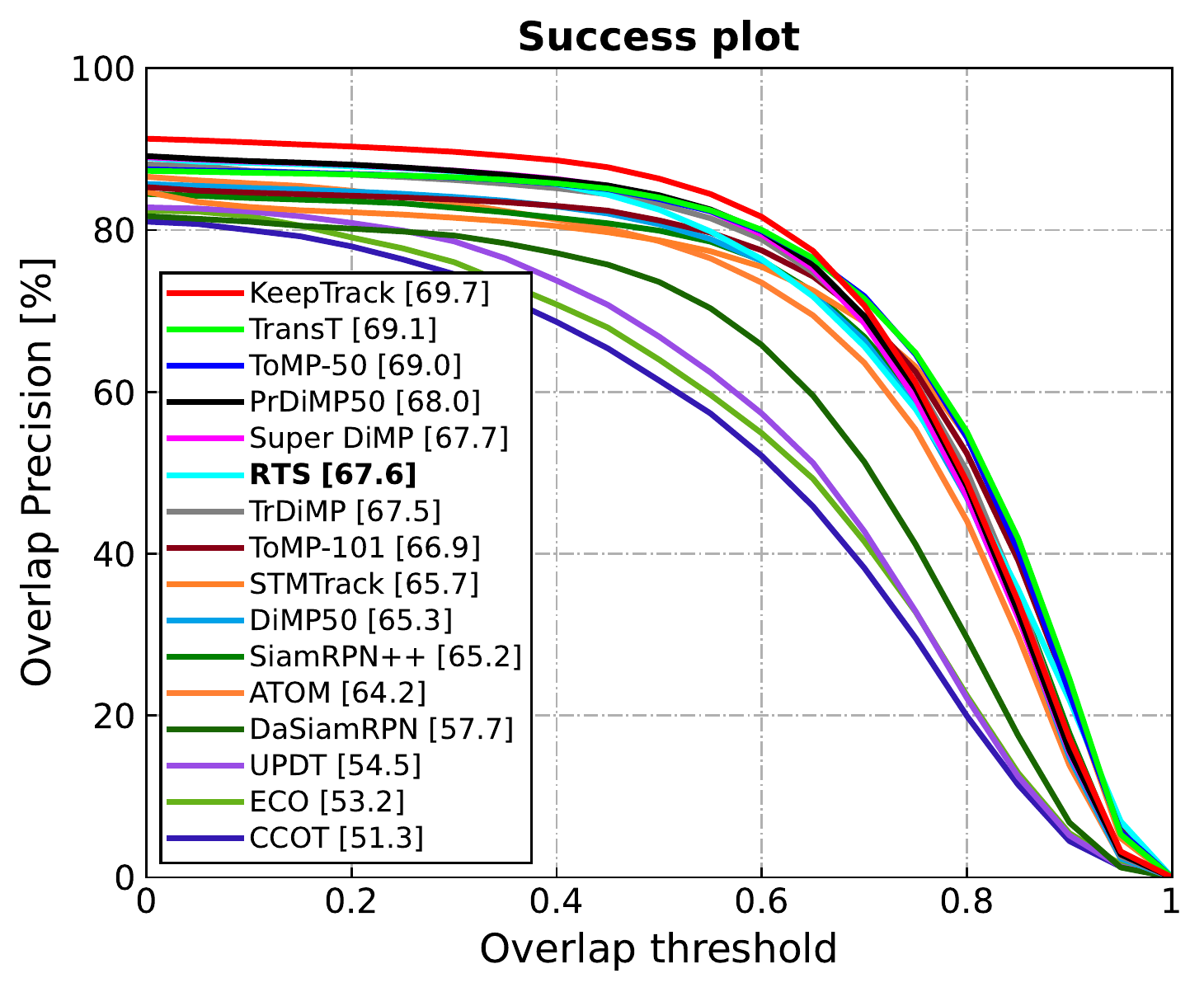}
    \includegraphics[width=0.45\linewidth, keepaspectratio]{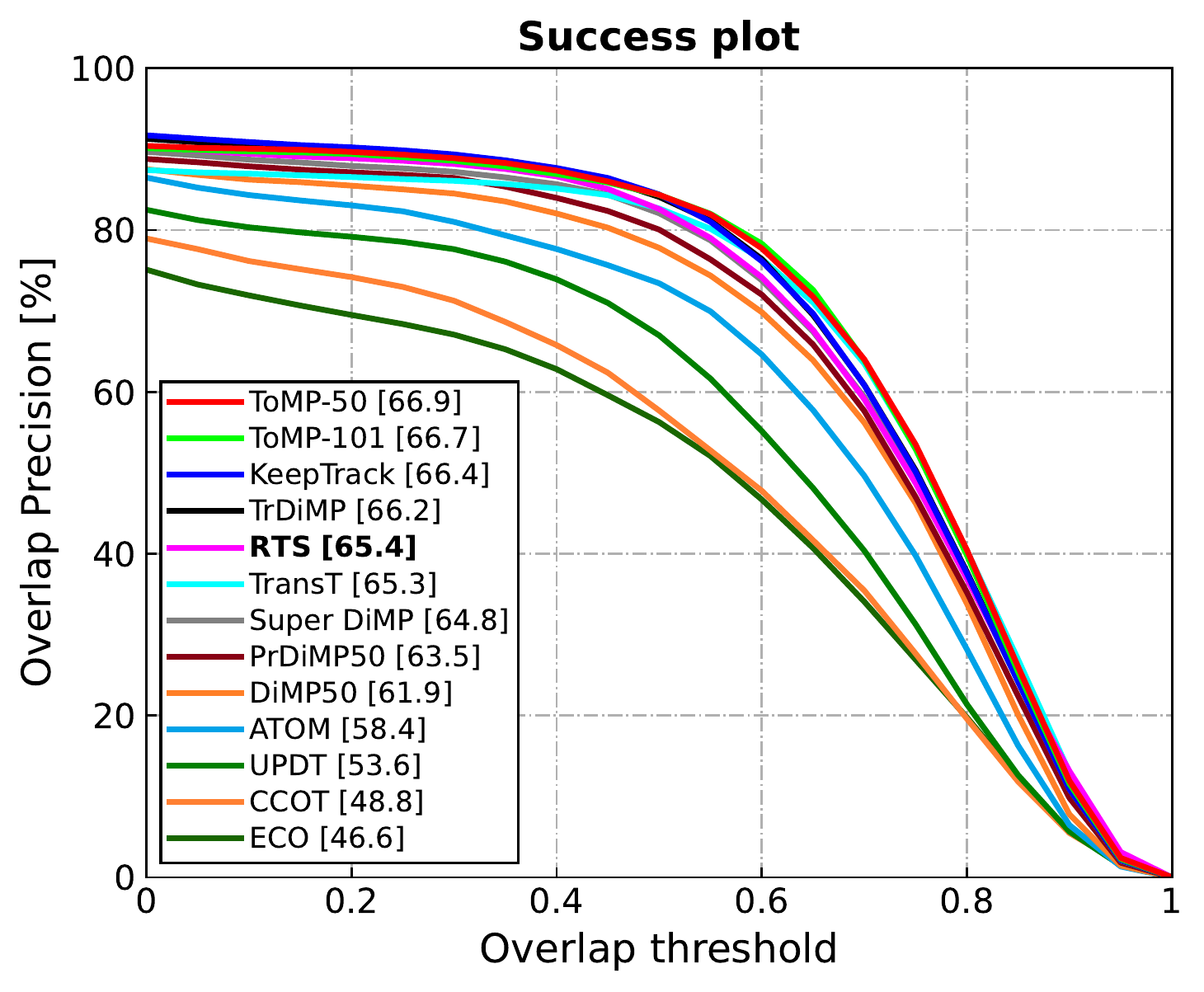}
\caption{Success plots on the UAV123~\cite{Mueller_2016_ECCV_UAV123} (left) and NFS~\cite{Galoogahi_2017_ICCV_NFS} (right) datasets in terms of overall AUC score, reported in the legend.}
\label{sup:fig:uav_nfs}
\end{figure}

\section{Additional Evaluation results}
\label{sup:sec:eval}

In this section we provide additional plots of our approach on different benchmarks, and a attribute analysis on LaSOT~\cite{LaSOT}.

\parsection{Success plots for LaSOT~\cite{LaSOT}, NFS~\cite{Galoogahi_2017_ICCV_NFS} and UAV123~\cite{Mueller_2016_ECCV_UAV123}}
\label{sup:sec:success_plots}
We provide in Figure~\ref{sup:fig:lasot_plots} all the plots for the metrics we report for LaSOT~\cite{LaSOT} in the paper: \emph{Success}, \emph{Normalized Precision} and \emph{Precision} plots. For completeness, we provide the success plots for NFS~\cite{Galoogahi_2017_ICCV_NFS} and UAV123~\cite{Mueller_2016_ECCV_UAV123} in Figure~\ref{sup:fig:uav_nfs}.

\begin{figure}[t]
\centering
\includegraphics[width=0.6\linewidth]{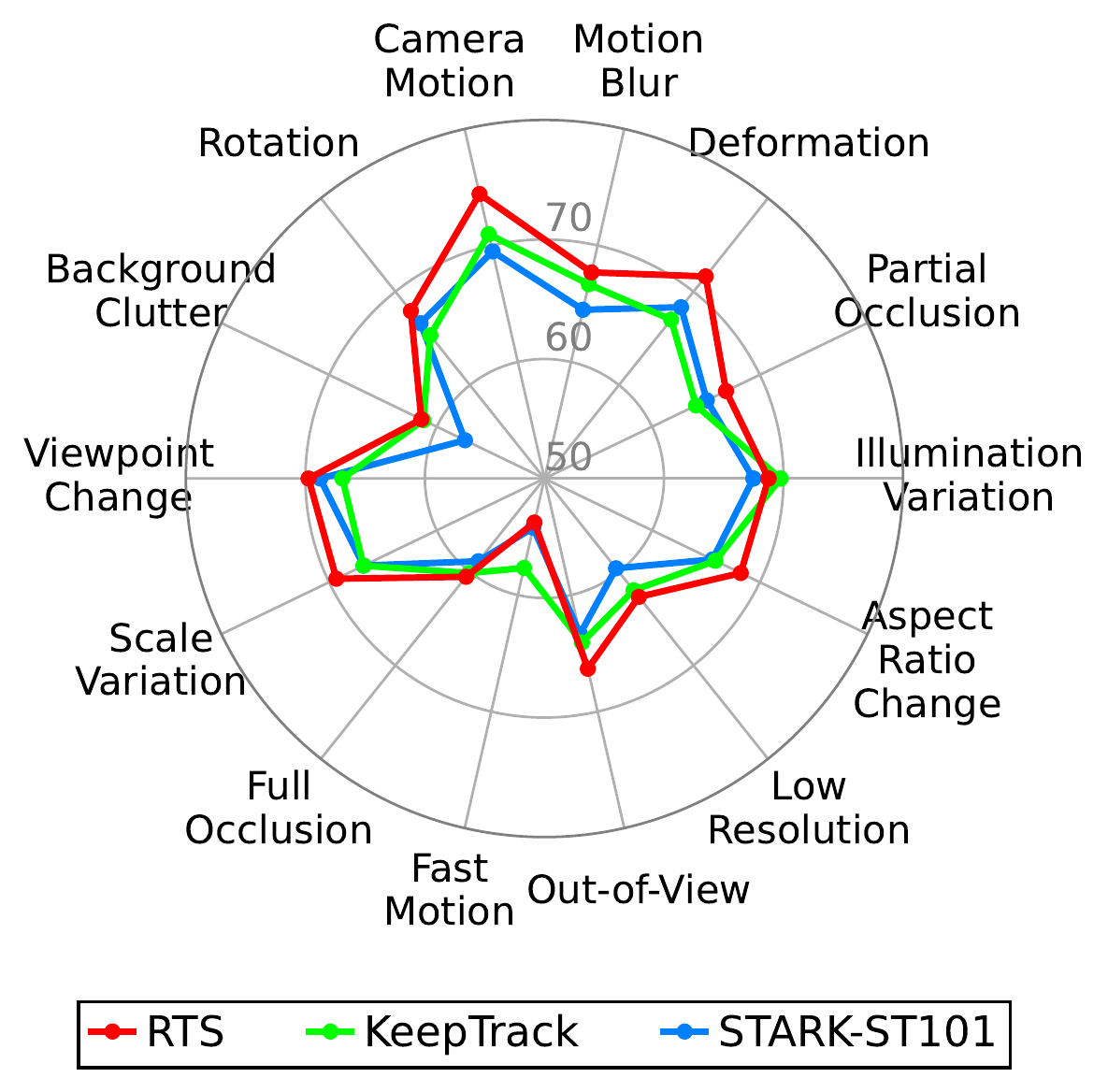}
\caption{Attributes comparison on LaSOT~\cite{LaSOT}.}%
\label{sup:fig:lasot_attributes_radar}
\end{figure}

\parsection{Attribute analysis on LaSOT~\cite{LaSOT}}
\label{sup:sec:attributes}
In this section, we focus on the dataset sequences attributes. We compare our approach to numerous other trackers, and provide the detailed results in Table~\ref{sup:tab:lasot_attributes}. Furthermore, we highlight the strength of our approach in Figure~\ref{sup:fig:lasot_attributes_radar} by focusing the comparison only to the two current state-of-the-art methods ToMP-101 and ToMP-50~\cite{Mayer_2022_CVPR_ToMP}. 

There are 14 attributes provided for LaSOT~\cite{LaSOT} sequences, representing different kind of challenges the tracker has to deal with in different situations. Compared to existing trackers, our method achieves better AUC scores in 11 out of 14 attributes. In particular, we outperform ToMP-50~\cite{Mayer_2022_CVPR_ToMP} and ToMP-101~\cite{Mayer_2022_CVPR_ToMP} by a large margin for the following attributes: \emph{Camera Motion} (+4.2\% and +2.7\%), \emph{Scale Variation} (+1.8\% and 0.9\%), \emph{Deformation} (+3.1\% and +2.2\%), \emph{Motion Blur} (+3.1\% and +2.5\%) and \emph{Aspect Ratio Change} (+1.7\% and +1.0\%). Our method is only outperformed on three attributes by KeepTrack~\cite{mayer_ICCV_2021_keeptrack} and ToMP~\cite{Mayer_2022_CVPR_ToMP} for \emph{Fast Motion} (-2.3\% to -4.1\%) and for \emph{Illumination Variation} (-0.3\% to -1.4\%). For \emph{Background Clutter}, RTS outperforms ToMP-50~\cite{Mayer_2022_CVPR_ToMP} by 2.3\% and fall just behind ToMP-101~\cite{Mayer_2022_CVPR_ToMP} (-0.1\%).

\begin{table*}[t]
	\caption{LaSOT~\cite{LaSOT} attribute-based analysis. Each column corresponds to the results computed on all sequences in the dataset with the corresponding attribute. Our method outperforms all others in 12 out of 14 attributes.}
	\centering
	\newcommand{\best}[1]{\textbf{\textcolor{red}{#1}}}
	\newcommand{\scnd}[1]{\textbf{\textcolor{blue}{#1}}}
	\newcommand{\dist}{\enspace}
	\resizebox{1.00\textwidth}{!}{%
        \begin{tabular}{l@{\dist}c@{\dist}c@{\dist}c@{\dist}c@{\dist}c@{\dist}c@{\dist}c@{\dist}c@{\dist}c@{\dist}c@{\dist}c@{\dist}c@{\dist}c@{\dist}c@{\dist}|c@{\dist}}
        	\toprule
        	& Illumination & Partial     &  \multirow{2}{*}{Deformation}              & Motion         & Camera      & \multirow{2}{*}{Rotation}            & Background & Viewpoint   & Scale        & Full        & Fast        &   \multirow{2}{*}{Out-of-View}          & Low         & Aspect        &  \multirow{2}{*}{Total}      \\
        	& Variation    & Occlusion   &     & Blur           & Motion      &     & Clutter    & Change      & Variation    & Occlusion   & Motion      &  & Resolution  & Ratio Change  &  \\
        	\midrule
            LTMU~\cite{Dai_2020_CVPR_LTMU}                     & 56.5         & 54.0        & 57.2           & 55.8           & 61.6        & 55.1        & 49.9        & 56.7        & 57.1        & 49.9        & 44.0        & 52.7        & 51.4        & 55.1          & 57.2 \\
            LWL~\cite{Bhat_2020_ECCV_LWL}                      & 65.3         & 56.4        & 61.6           & 59.1           & 64.7        & 57.4        & 53.1        & 58.1        & 59.3        & 48.7        & 46.5        & 51.5        & 48.7        & 57.9          & 59.7 \\
            PrDiMP50~\cite{danelljan_ICCV_2019_PRDimP}         & 63.7         & 56.9        & 60.8           & 57.9           & 64.2        & 58.1        & 54.3        & 59.2        & 59.4        & 51.3        & 48.4        & 55.3        & 53.5        & 58.6          & 59.8 \\
            STMTrack~\cite{Fu_2021_CVPR_STMTrack}              & 65.2         & 57.1        & 64.0           & 55.3           & 63.3        & 60.1        & 54.1        & 58.2        & 60.6        & 47.8        & 42.4        & 51.9        & 50.3        & 58.8          & 60.6 \\
            SuperDiMP~\cite{bhat_ICCV_2019_DiMP}               & 67.8         & 59.7        & 63.4           & 62.0           & 68.0        & 61.4        & 57.3        & 63.4        & 62.9        & 54.1        & 50.7        & 59.0        & 56.4        & 61.6          & 63.1 \\
            TrDiMP~\cite{Wang_2021_CVPR_TrDiMP}                & 67.5         & 61.1        & 64.4           & 62.4           & 68.1        & 62.4        & 58.9        & 62.8        & 63.4        & 56.4        & 53.0        & 60.7        & 58.1        & 62.3          & 63.9 \\
            Siam R-CNN~\cite{Voigtlaender_2020_CVPR_SiamRCNN}  & 64.6         & 62.2        & 65.2           & 63.1           & 68.2        & 64.1        & 54.2        & 65.3        & 64.5        & 55.3        & 51.5        & 62.2        & 57.1        & 63.4          & 64.8 \\
            TransT~\cite{Chen_2021_CVPR_TransT}                & 65.2         & 62.0        & 67.0           & 63.0           & 67.2        & 64.3        & 57.9        & 61.7        & 64.6        & 55.3        & 51.0        & 58.2        & 56.4        & 63.2          & 64.9 \\
            AlphaRefine~\cite{Yan2021AlphaRefineBT}            & 69.4         & 62.3        & 66.3           & 65.2           & 70.0        & 63.9        & 58.8        & 63.1        & 65.4        & 57.4        & 53.6        & 61.1        & 58.6        & 64.1          & 65.3 \\
            KeepTrack Fast~\cite{mayer_ICCV_2021_keeptrack}    & \best{70.1}  & 63.8        & 66.2           & 65.0           & 70.7        & 65.1        & 60.1        & 67.6        & 66.6        & 59.2        & 57.1        & 63.4        & 62.0        & 65.6          & 66.8 \\
            KeepTrack~\cite{mayer_ICCV_2021_keeptrack}         & \scnd{69.7}  & 64.1        & 67.0           & \scnd{66.7}    & 71.0        & 65.3        & 61.2        & 66.9        & 66.8        & \scnd{60.1} & \scnd{57.7} & \scnd{64.1} & 62.0        & 65.9          & 67.1 \\
            STARK-ST101~\cite{Yan_2021_ICCV_STARK}             &       67.5   & 65.1        & 68.3           & 64.5           & 69.5        & 66.6        & 57.4        & 68.8        & 66.8        & 58.9        & 54.2        & 63.3        & 59.6        & 65.6          & 67.1 \\
            ToMP-50~\cite{Mayer_2022_CVPR_ToMP}       & 66.8         & 64.9        & 68.5           & 64.6           & 70.2        & 67.3        & 59.1        & 67.2        & 67.5        & 59.3        & 56.1        & 63.7        & 61.1        & 66.5          & 67.6 \\
            ToMP-101~\cite{Mayer_2022_CVPR_ToMP}      & 69.0         & \scnd{65.3} & \scnd{69.4}    & 65.2           & \scnd{71.7} & \scnd{67.8} & \best{61.5} & \scnd{69.2} & \scnd{68.4} & 59.1        & \best{57.9} & \scnd{64.1} & \scnd{62.5} & \scnd{67.2}   & \scnd{68.5} \\
            \textbf{RTS}                                       &        68.7  & \best{66.9} & \best{71.6}    & \best{67.7}    & \best{74.4} & \best{67.9} & \scnd{61.4} & \best{69.7} & \best{69.3} & \best{60.5} & 53.8        & \best{66.3} & \best{62.7} & \best{68.2}   & \best{69.7} \\
            \bottomrule
        \end{tabular}
        
	}
	\label{sup:tab:lasot_attributes}
\end{table*}

\section{Additional Content}
\label{sup:sec:morecontent}

\begin{figure*}[!b]
\centering
\includegraphics[width=1.0\linewidth]{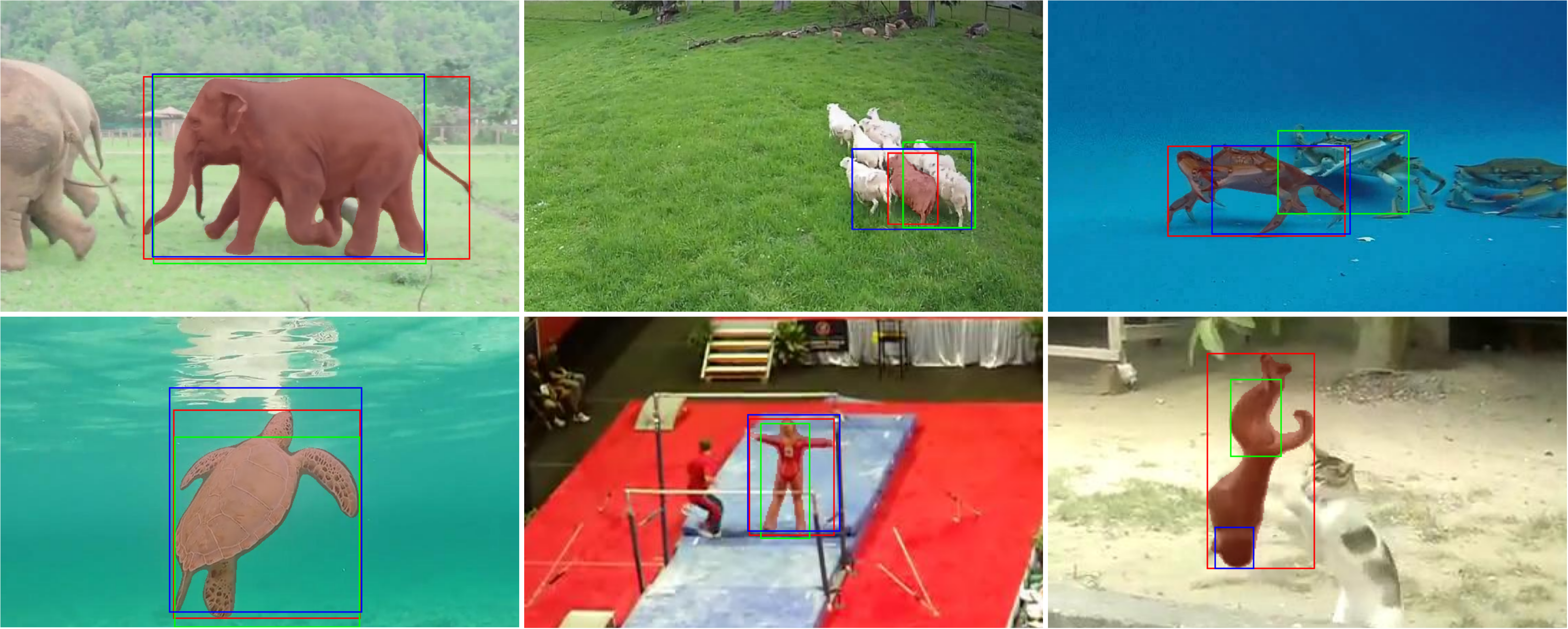}
\caption{Qualitative results on LaSOT~\cite{LaSOT} of our approach compared to the previous state-of-the-art methods KeepTrack~\cite{mayer_ICCV_2021_keeptrack} and STARK-ST101~\cite{Yan_2021_ICCV_STARK}. As they do not produce segmentation masks, we represent ours as a red overlay and print for all methods the predicted bounding boxes with the following color code:\\ \bbox{_green}~KeepTrack~~~~\bbox{_blue}~STARK-ST101~~~~\bbox{_red}~RTS}%
\label{fig:qualitative_lasot}
\end{figure*}

Figure~\ref{fig:qualitative_lasot} shows additional visual results compared to other state-of-the-art trackers on 6 different sequences of LaSOT~\cite{LaSOT}. For more content, we refer the reader to: \url{https://github.com/visionml/pytracking}.

\end{document}